\documentclass{article}

    \PassOptionsToPackage{numbers, compress}{natbib}

    \usepackage[preprint]{neurips_data_2022}

\usepackage[utf8]{inputenc} 
\usepackage{float}
\usepackage[T1]{fontenc}    
\usepackage{hyperref}       
\usepackage{url}            
\usepackage{booktabs}       
\usepackage{amsfonts}       
\usepackage{nicefrac}       
\usepackage{microtype}      
\usepackage{amsmath, amssymb}
\usepackage{mathtools}
\usepackage{algorithm}
\usepackage[noend]{algpseudocode}
\usepackage{lipsum}
\usepackage{multirow}
\usepackage{makecell}
\usepackage[dvipsnames]{xcolor}
\PassOptionsToPackage{usenames,dvipsnames}{xcolor}
\usepackage{cleveref}
\newcommand\myshade{85}

\colorlet{mylinkcolor}{violet}
\colorlet{mycitecolor}{YellowOrange}
\colorlet{myurlcolor}{Aquamarine}
\DeclarePairedDelimiter\ceil{\lceil}{\rceil}

\hypersetup{
  linkcolor  = mylinkcolor!\myshade!black,
  citecolor  = mycitecolor!\myshade!black,
  urlcolor   = myurlcolor!\myshade!black,
  colorlinks = true,
}

\title{A Framework for Large Scale Synthetic Graph Dataset Generation}

\author{%
  Sajad Darabi \thanks{equal contribution} \\
  \texttt{sdarabi@nvidia.com} \\
   \And
   Piotr Bigaj \footnotemark[1] \\
   \texttt{pbigaj@nvidia.com} \\
   \AND
   Dawid Majchrowski \\
   \texttt{dmajchrowski@nvidia.com} \\
   \And
   Artur Kasymov \\
\texttt{akasymov@nvidia.com} \\
    \AND
   Pawel Morkisz \\
   \texttt{pmorkisz@nvidia.com} \\
   \And
   Alex Fit-Florea \\
   \texttt{afitflorea@nvidia.com} \\
}
\begin{document}

\maketitle

\begin{abstract}
Recently there has been increasing interest in developing and deploying deep graph learning algorithms for many tasks, such as fraud detection and recommender systems. Albeit, there is a limited number of publicly available graph-structured datasets, most of which are tiny compared to production-sized applications or are limited in their application domain. This work tackles this shortcoming by proposing a scalable synthetic graph generation tool to scale the datasets to production-size graphs with trillions of edges and billions of nodes. The tool learns a series of parametric models from proprietary datasets that can be released to researchers to study various graph methods on the synthetic data increasing prototype development and novel applications. We demonstrate the generalizability of the framework across a series of datasets, mimicking structural and feature distributions as well as the ability to scale them across varying sizes demonstrating their usefulness for benchmarking and model development. Code can be found on \href{https://github.com/NVIDIA/DeepLearningExamples/tree/master/Tools/DGLPyTorch/SyntheticGraphGeneration}{github}.
\end{abstract}

\section{Introduction}

Graphs are widely used data structures that capture relational and structural information between individual entities (nodes) via connections (edges) in many domains. For example, in social networks, a graph-based learning system leverages structural and feature information to make accurate user recommendations. Similarly, in an e-commerce platform, a transaction network can be used to detect fraudulent transactions. Real-world graphs are diverse. For a simple recommendation scenario consisting of user and item nodes, the user nodes would include information about age, gender, and income. Whereas the item nodes (e.g. a movie) would be characterized by the genre, length, and list of actors. Additionally, edge features may contain information about the rating the user gave a movie. Such attributed graphs are prevalent, where the graph dataset's structure is enriched with features of the nodes and edges.

Graph neural networks (GNNs) have recently received increasing attention due to the wide range of applications that deal with data naturally represented as graphs. Motivated by similar developments in other domains, there have been efforts to extend the benefit of deep learning to this non-Euclidean domain enabling more streamlined approaches that leverage the relational data. Various methods have been developed to learn from graph data, such as Node2Vec \citep{grover2016node2vec}, graph convolution networks (GCN) \citep{kipf2016semi}, and graph attention networks (GAT) \citep{velivckovic2017graph} which have been used for a variety of tasks including node classification \citep{kipf2016semi}, link prediction \citep{liben2007link}, graph clustering \citep{hamilton2017representation}. These methods are collectively referred to as geometric deep learning \cite{bronstein2017geometric}. 

A central problem in geometric deep learning is the need for real-world datasets that are large enough for industry-scale problems. Most of the larger public datasets are similar and are often derived from academic citation networks \cite{hu2021ogb}, which are too small for these problems. This lack of diversity limits the development of graph neural networks (GNN) and their evaluation. In this work, we propose a \textit{framework for synthetic graph generation} which can systematically generate graphs with corresponding node or edge features in a scalable manner. Generating realistic large-scale graph datasets, which we define as graphs with billions to trillions of edges that simulate real-world datasets distributions, will enable data sharing and facilitate the development of GNNs that scale to such large graph size. This dataset curation will be a key component to advancing the field both for developing models that scale and from the perspective of improving the accuracy of developed GNNs, and new efficient geometric deep learning methods.
 
The proposed framework provides a parametric model that is flexible enough to fit a single graph as well as many graphs. Our main contributions are as follows, 
\begin{enumerate}
     \item We propose a framework for synthetic graph dataset generation that can generate a graph of arbitrary size based on original (usually smaller) graph characteristics containing both structure and the associated node/edge features. 
     \item We show a set of case studies reflecting the generality and effectiveness of our approach on real-world datasets, simulating real-world graph statistical properties. 
 \end{enumerate}

The paper is organized as follows: in the next sections we provide an overview of recent methods, then we define the problem and proposed method,  subsequently the experimental setup to evaluate the method is presented, and finally a result section showcasing the usability of such framework on real-world datasets.

\section{Related Work}



There has been increasing interest in curating  datasets for different graph prediction tasks; for example, for node classification (CORA, PUBMED, and CITESEER)  are commonly used \cite{kipf2016semi}, for link prediction (WN18, FB15k, OAG) \cite{schlichtkrull2018modeling, qiu2019netsmf}. More recently, Open Graph Benchmark (OGB) datasets have been used for a set of challenging and realistic datasets to facilitate graph machine learning research and application development \cite{hu2020open}. However, most of these sources are limited in scope as they are primarily derived from citation networks or social networks, limiting the scientific insight into various problems that derive similar graph data representations. As a result, synthetic graph generation has been proposed as a facilitator to this gap for investigating different models in this domain. 

The development of generative models for graphs poses unique challenges. These generative models are broadly categorized into two categories: traditional model-based methods and deep learning-based methods. Simple, elegant, and general mathematical models are instrumental in graph generation. The two simplest random graph models are to select a graph uniformly at random commonly referred to as the Erdős–Rényi models \cite{erdos59a}. In biology, these models are accepted as a basic model to study biological networks where the similarity of typologies and biological regulatory networks are investigated or compared against \cite{ma2008proximity}. Despite its usability and scalability, it violates power laws commonly found in social networks. A class of procedural generators tries to find simple mechanisms to generate graphs that match this property of graphs commonly found in the real-world. A typical representative here is the Barabási–Albert Method \cite{albert2002statistical,PhysRevLett.85.5234}, which uses a preferential attachment idea where new nodes prefer to connect to existing nodes. R-MAT is another example of such a random graph generation model \citet{chakrabarti2004r}. An alternative well-studied random graph model is the Stochastic Block Model \citep{abbe2017community} which generates graphs based on communities within the graph and their degree distributions. Similarly, the Stochastic Kronecker Graph (SKG) method introduced in \cite{10.1007/11564126_17} is a generalization of R-MAT for finding power-law degree distribution. The above methods are very attractive from a computational complexity standpoint, as well as modeling real-world graph properties.

On the other hand, recent deep-graph generators (DGG) that generate graphs sequentially either by generating node-by-node, edge-by-edge, or a block of nodes have been proposed. These generators are autoregressive \citep{you2018graphrnn, liao2019gran, martinkus2022spectre} and are limited in generating small graphs with 100s-1000s nodes. Further, very few deep graph generators work with single graph inputs, commonly referred to as one-shot graph generators, which are auto-encoder based \cite{kipf2016variational, grover2019graphite}. Lastly, most DGG methods lack the ability to generate features in addition to the structure, primarily because of the complexity of modeling such a problem jointly in an end-to-end fashion.  A good overview of DGG can be found in \cite{dgg_survy}. Similarly, recent deep-learning methods such as variational autoencoders (VAE) \cite{kingma2013auto}, and generative adversarial networks (GAN) \cite{goodfellow2014generative} have overtaken traditional tabular generators such as kernel density estimators (KDE) \cite{rosenblatt1956remarks, parzen1962estimation}. For a more comprehensive review on tabular methods refer to \cite{borisov2022deep}.


\section{Proposed Method}
\subsection{Problem Formulation}
Our objective is to propose a graph generation framework that supports trillions of edges/nodes, along with node and edge features. We describe an example of how to use the framework to generate large graphs from single graph input graphs. The framework is flexible and the components can be swapped with other alternatives, and is not limited to the specific components described. The limitations of the framework are discussed in Section \ref{sec:broaderimpact}.

Formally, a graph contains both structural information and features; as such we define the graph as a triple $G(S, F_{\mathcal{V}}, F_{\mathcal{E}})$, where $S = (\mathcal{V}, \mathcal{E})$
, $\mathcal{V} = \{v_1, v_2, \cdots, v_\mathcal{N}\}$ is the set of $\mathcal{N}$ nodes (or vertices), $F_{\mathcal{V}} \in \Re^{\mathcal{N} \times d_{\mathcal{V}}}$ is the corresponding feature matrix associated with node features with dimension $d_{\mathcal{V}}$, $\mathcal{E} \subset \mathcal{V}\times\mathcal{V}$ is the set of $\mathcal{M}$ edges, where $e_{ij}$ is an edge that connects node $v_i$ and $v_j \in \mathcal{V}$, and $F_{\mathcal{E}} \in \Re^{\mathcal{M} \times d_{\mathcal{E}}}$ is the associated edge feature for each edge $e_{ij} \in \mathcal{E}$ and $d_{\mathcal{E}}$ is the dimension of each edge feature.

Given an input graph $G$ with an arbitrary number of nodes, edges, and feature sets we aim to learn the probabilistic model that generated this graph $p(G)$. New graphs are then sampled from this generative process $\tilde{G} \sim p_{model}(G)$. In a general setting, a graph with $\mathcal{N}$ nodes can be represented by up to $\mathcal{N}!$ adjacency matrices $A^{\pi}$, corresponding to arbitrary node ordering, resulting in a high representation complexity, especially for large graphs. In the undirected case, there are $2^{\mathcal{N}(\mathcal{N}-1)/2}$ of such graphs.  It is important for generative models to scale to large-scale (billions or more edges/nodes) graphs and to accommodate this complexity in the output space. Additionally, simultaneously generating node features $F_{\mathcal{V}}$ and edge features $F_{\mathcal{E}}$ greatly increases this modeling complexity. To make the problem tractable we decompose the generative process into different components

\begin{figure}
    \centering
    \includegraphics[scale=1.0]{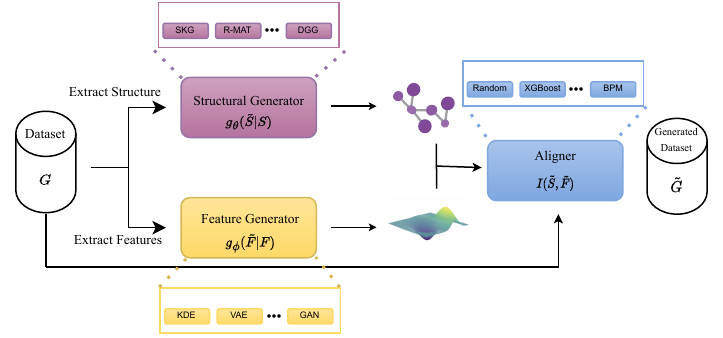}
    \caption{Overview of our proposed synthetic data generation framework. The module is composed of three parts: a structural generator, which fits the graph structure, a feature generator which fits the feature distribution contained in the graph, and finally an aligner, which aligns the generated features with the generated graph structure.}
    \label{fig:overview}
\end{figure}

Our proposed generative model consists of three components: structural generation, feature generation, and alignment of these components as depicted in Figure \ref{fig:overview}. As shown, we make the structural generation and feature generation independent, which are then brought together using an aligner  $\mathcal{A}(S, F) \rightarrow \tilde{G}$ (the aligner is detailed in section \ref{sec:aligner}). In the following sections we will detail each component within the framework. 

Here on, to simplify the notation we consider a single graph $G$ as input to the generator. From this graph we extract the corresponding structural information $S$, and its associated feature sets $F_\mathcal{V}$, $F_\mathcal{E}$. Next, we detail the structural generator $g_\theta$.

\subsection{Structure Generation}
\subsubsection{Motivation}
As we are primarily interested in scaling the generation tool to trillion edge graphs, and supporting single graph input datasets we leverage model-based graph generators. The model can be seen as a generalized stochastic Kronecker matrix multiplication. The graph structure consists of the corresponding nodes and edges without node or feature attributes, i.e. $S = (\mathcal{V}, \mathcal{E})$. The adjacency matrix $A$ corresponding to this graph is an $n \times m$ matrix, where $\mathcal{N} = n + m$, with entries $a(i, j) = 1$ if the edge $e_{ij}$ between node $i$ and node $j$ exists. Note an $n \times n$ matrix with constraints can be used as an identical representations but will carry redundancies and complications in the generation scheme (e.g. 0's in bipartite graphs, varied degree distribution between partites, etc).

We will first formalize the problem of generating the structure for a simple graph $G$ that is non-directed heterogeneous in nodes and homogeneous in edges. Later we will extend this to general graph to show that our generator is a generalisation of R-MAT \cite{rmat}.

\subsubsection{Problem formulation}
The objective is to generate a graph $\tilde{G}$ represented by the adjacency matrix $\tilde{A}$ that is similar to the original graph adjacency matrix $A$. The $\tilde{A}$ adjacency is generated by sampling $E$ edges from a distribution $\theta$. Here $\theta$ is a discrete 2D probability distribution, where $\theta_{i,j}$ represents the probability a directed edge from node $i$ to node $j$ exists in graph $\tilde{G}$. 
If we choose to generate the graph of the same size as the input, that is $\tilde{A}$ is $n \times m$ and graph $\tilde{G}$ has $n+m$ nodes, and the original graph $A$ has $N + M$ nodes then $\tilde{A} \sim \theta$ which is generated as follows
\begin{equation} \label{eq:theta_matrix_eq}
 \theta = \theta_S^{\otimes min(m,n)} \otimes  \theta_H^{\otimes max(0,n-m)} \otimes  \theta_V^{\otimes max(0,m-n)}
 \end{equation}
 where
\begin{equation} 
    \theta_S =  \begin{bmatrix}
                a & b \\
                c & d 
            \end{bmatrix}, 
    ~\theta_H =  \begin{bmatrix}
                q & 1-q 
            \end{bmatrix},
    ~\theta_V =  \begin{bmatrix}
                p \\ 
                1-p 
            \end{bmatrix},
    \linebreak
\end{equation}
\vspace{-3.0mm}
\begin{equation}\label{eq:mn_equation}
   m=\ceil{\log_2 M} ,\quad n=\ceil{\log_2 N},
\end{equation}
\vspace{-4.0mm}
\begin{equation}\label{eq:mn_equation_pq}
    p=a+b, \quad q=a+c,
\end{equation}
$\otimes$ is a Kronecker matrix product, and $A^{\otimes b}=\underbrace{A \otimes A \otimes ... \otimes A}_{b times}$ is the matrix Kronecker power.
Here $\theta_H$ and $\theta_V$ are marginals of $\theta_S$, where $a, b, c, d$ is the probability of an edge occurring in the partition and $a+b+c+d=1$. The marginals depends only on the shape of matrix $\tilde{A}$ and only one of them is effectively used to form the probability distribution $\theta$. If matrix  $\tilde{A}$ is square ($n=m$) then
\begin{align}
    \theta = \theta_S^{\otimes n}
\end{align}
which effectively is an R-MAT algorithm. The proposed approach differs in comparison to previous methods as $\tilde{A}$ might be a non-square adjacency matrix, additionally $\tilde{A}$ in general is constructed in a way that the $i$-th row and the $i$-th column can represent different nodes in the graph, giving us the ability to construct graphs that are heterogeneous in nodes in K-partite graphs. To scale this further and generate graphs that don't fit on system memory, we use a chunked generation algorithm detailed in Appendix \ref{sec:chunked_generation}.

For the K-partite graph, the obtained adjacency matrix is a block matrix with connectivity between the nodes in the corresponding partite. In this framework in order to represent K-partite graphs with $\tilde{A}$ where matrix coordinates imply node ids then, it is sufficient to consider $\tilde{A}_P$ for each partite $P$ and create $\tilde{A}$ out of $\tilde{A}_P$ for each partite. 

\subsubsection{Fitting the generator}

The purpose of the structure generator is to generate a graph $\tilde{G}$ that is similar in graph characteristics to the original graph $G$. These characteristics will depend on the parameters of $\theta_S$. As similarity measure we use the normalized degree distributions of the graphs $G$, and $\tilde{G}$. This metric is widely used to estimate other properties of the network structure and allows to build general models for complex network connectivity patterns \cite{albert2002statistical, wills2020metrics}. In order to find parameters of $\theta_S$ we need to minimize the following error
\begin{equation} \label{eq:gstructure_generator_loss}
    J(\theta_S) \propto \sum_{k^{in}=0}^{k_{max}^{in}} (c_k^{in} - \tilde{c_k^{in}})^2 + \sum_{k^{out}=0}^{k_{max}^{out}} (c_k^{out} - \tilde{c_k^{out}})^2,
\end{equation}
where $k$ is the node degree, $c_{k}^{in}$ is the number of nodes in graph $G$ having in-degree $k$, $c_{k}^{out}$ is the number of nodes in graph $G$ having out-degree $k$,
$\tilde{c_{k}^{in}}$ is the estimated number of nodes in graph $\tilde{G}$ having in-degree k, and $\tilde{c_{k}^{out}}$ is the estimated number of nodes in graph $\tilde{G}$ having out-degree k.

Note that this does not imply that $A$ and $\tilde{A}$ are similar in the sense $A -\tilde{A} \approx 0$, which would require appropriate node permutation in  $\tilde{G}$. Simple degree distributions comparison allows us to effectively compare graphs only when the number of nodes is equivalent, though our goal is to generate a graph $\tilde{G}$ that is arbitrarily larger in number of nodes than $G$. Therefore, we require a $\theta_S$ for the graph of the same size and then generate $\tilde{c_{k}^{in}}$ and $\tilde{c_{k}^{out}}$ as follows

\begin{equation} \label{eq:ckout_eq}
    \tilde{c_{k}^{out}} = {E\choose k} \sum_{i=0}^{m} {m\choose i} [p^{m-i}(1-p)^i]^k[1-(p^{m-i}(1-p)^i)]^{E-k} 
\end{equation}
and
\begin{equation} \label{eq:ckin_eq}
    \tilde{c_{k}^{in}} = {E\choose k} \sum_{i=0}^{n} {n\choose i} [q^{n-i}(1-q)^i]^k[1-(q^{n-i}(1-q)^i)]^{E-k}, 
\end{equation}

where $\tilde{c_{k}^{in}}$ and $\tilde{c_{k}^{out}}$ depend only on $p$ and $q$. Solving \eqref{eq:gstructure_generator_loss} for $p=a+b$ and $q=a+c$ leads to underdetermined system ($a+b+d+c=1$) as we have 3 equations and 4 variables. \cite{rmat} proposes to use $\frac{a}{b}=\frac{a}{c}=\frac{3}{1}$, since this can be seen  in  many real-world scenarios. We have encountered graph datasets that did not follow this ratio. Instead of this ratio, we propose to estimate  $\frac{a}{b}$ and $\frac{a}{c}$ from adjacency matrix $A$ of the original graph $G$ by Maximum Likelihood Estimation of these parameters.

\subsection{Feature Generation}
\label{sec:feature_gen}
Next we consider the feature sets associated with the graph $\mathcal{D}_{\textrm{features}} = F_{\mathcal{V}/\mathcal{E}}$, where each row $x_i \in \Re^{d_{\mathcal{V}/\mathcal{E}}}$ is an observation sampled from a data-generating distribution $P_{F_{\mathcal{V}/\mathcal{E}}}(x)$. We treat this dataset as a tabular dataset. Each row of the data set contains the edge features, source node features, and destination node features. The objective is to learn a generative model over this data generating process. To this end, we consider the multi-modal setting where $x_i$ is a concatenation of discrete $\mathcal{D} = [D_1, \cdots, D_{|\mathcal{D}|}]$ and continuous features $\mathcal{C} = [C_1, \cdots, C_{|\mathcal{C}|}]$. Without loss of generality, our generator follows a GAN architecture, though any high-capacity method that can model the underlying distribution can be used. To this end, our input layer consists of a feature tokenizer where the corresponding embedding for each feature is computed as follows:
\begin{align}
    E_j = b_j + f_j(x_j) \in \Re^{d_j} \quad f_j: \mathbb{X} \rightarrow \Re^{d_j}
\end{align}

where $b_j$ is a bias term for the $j\textrm{-th}$ feature and $f_j$ is the corresponding feature tokenization function. Our model applies mode-specific normalization to continuous columns which 1) fits a Variational Gaussian Mixture (VGM) to continuous columns of $\mathcal{C}$ 2) converts the continuous elements of $\mathcal{C}_i$ into a one-hot vector denoting the specific Gaussian that best matches the element as well as its scalar value normalized within the selected Gaussian as done in \citep{NEURIPS2019_254ed7d2}. For discrete columns we introduce embedding layers $W_{D_i} \in \Re^{|D_i| \times d_{D_i}}$, where $|D_i|$ is the number of possible unique discrete values and $d_{D_i}$ is the dimension of the embedding. The input layer operation could be summarized as follows
\begin{alignat}{3}
    &E_j^{cont} &&= f_j^{cont}(C_j) && \in \Re^{d_{C_j}}\\
    &E_j^{cat}  &&= b_j^{cat} + e(D_j)^TW_j^{cat} && \in \Re^{d_{D_j}} \\
    &\bar{X}          &&= \textit{concat}[E_1^{cont}, \cdots, E_{C_{|\mathcal{C}|}}^{cont}, E_1^{cat}, \cdots, E_{D_{|\mathcal{D}|}}^{cat}] && \in \Re^{d_{\bar{X}}},
\end{alignat}
where $f_j^{cont}$ is a single layer fully-connected network, $e(\cdot)$ converts the input into a one hot vector, and the tokenized input $\bar{X}$ has a dimension $d_{\bar{X}} = \sum_{j}^{|\mathcal{C}|}d_{C_j} + \sum_{j}^{|\mathcal{D}|}d_{D_j}$. In GAN training there are two separate models the generator $\mathcal{G}$ and discriminator $D$ that estimates the probability of whether the sample came from the fake or real distribution. The generator $\mathcal{G}: \Re^{dim(z)} \rightarrow \Re^{dim(x)}$ takes the input $z \sim p(z) \in \Re^{d_z}$ and recovers samples in the original data space $\tilde{x}$. The discriminator then distinguishes between $D(x, \tilde{x}) \rightarrow [0, 1]$. Both networks are high capacity deep neural networks that follow a stack of $f(x) = \theta(\texttt{ResNetBlock}(\cdots (\texttt{ResNetBlock(FC(x))))})$ where $\texttt{ResNetBlock}(x) = x + \texttt{Dropout(ReLU(FC(BatchNorm(x))))}$, $\texttt{FC}$ is a fully-connected network which takes the $dim(x)$ size of its input. The networks $\mathcal{G}$ and $D$ are both trained together using the GAN objective 
\begin{align}
    \min_\mathcal{G}\max_D l(\mathcal{G}, D) = \min_\mathcal{G}\max_D &\mathbb{E}_{x\sim p_{data}(x)}[log(D(x)] \\ + &\mathbb{E}_{z\sim p(z)}[log(1 - D(\mathcal{G}(z)))].
\end{align}
The trained generator $\mathcal{G}$ is then used to sample feature sets $\tilde{F}_{\mathcal{V}/\mathcal{E}} = \mathcal{G}(z)$ for $z \sim p(z)$.

\subsection{Aligner}
\label{sec:aligner}
Once the structural and feature generators are trained the final graph is created via an aligner. The aligner is a function that maps the generated set of features onto the generated graph structure $\mathcal{I}(S, F) \rightarrow G(\mathcal{V}, \mathcal{E}, F_\mathcal{V}, F_\mathcal{E})$. A trivial aligner could randomly assign features to the corresponding nodes and edges of the generated graph. Instead, we propose to train a function $R$ that matches the generated structure with the generated features, preserving some properties of the input graph $G$. For example, for a recommender system use case, we may have an advertisement that is clicked by the majority of the population and we want to preserve its features properties.

Given the real graph $G_{real}$ we extract a set of features from the graph $F_S: V \rightarrow \Re^{d_{S}}$. These features correspond purely with the graph structure, such as node degree, node centrality, clustering coefficient, and page rank. More sophisticated features can also be considered such as using a GNN to obtain embeddings (refer to Section \ref{sec:aligncomparison}). Subsequently, a predictor $R$ is trained to capture the correlation between the graph structural features and the corresponding feature sets. For an edge $e_{(src, dst)}$ the predictor  $R : \Re^{d_{S}} \times \Re^{d_{S}} \rightarrow \Re^{d_\mathcal{E}}$ maps it on to the feature $x=R(F_S(v_{src}), F_S(v_{dst}))$, where $src$ is the source node index, and $dst$ is the destination node index.  We choose $\texttt{XGBoost}$ \cite{chen2016xgboost} as our predictor $R$, for each feature $x_j$. The series of $\texttt{XGBoost}$ models are trained to infer the features from structural information. For aligning the edges $e_{src, dst}$ both features of node $F_S(v_{src})$ and $F_S(v_{dst})$ are used as input to the model. The proposed aligner can be summarized as follows
\begin{alignat} {1}
    R(F_S(\mathcal{E}_{src}), F_S(\mathcal{E}_{dst})) &= \texttt{concat}[\texttt{XGBoost}_1, \cdots, \texttt{XGBoost}_k] \in \Re^{d_\mathcal{E}}\\
   &= [\hat{x}_1, \cdots, \hat{x}_k] = \hat{X} \\
    \textrm{Rank} &= \max_{i\in M}(\texttt{sim}(\hat{X}, X_i)),
\end{alignat}
where $\mathcal{E}_{src}$ and $\mathcal{E}_{dst}$ are the set of source and destination vertices, respectively, $\texttt{concat}$ is the concatenation operation, and $\texttt{sim} : \Re^{d} \times \Re^{d} \rightarrow \Re^+$ measures the similarity between the inputs and will be described below. 

The similarity score between the predicted vector and the corresponding generated feature is used to rank the features that are assigned to the edges of the graph. For continuous values, the mean squared error is used 
\begin{align}
    -\sum_{j \in \mathcal{C}} (R(F_S(v_{src}), F_S(v_{dst}))^{(j)} - x^{(j)}_i)^2
\end{align}
and similarly for categorical columns the cosine similarity is used
\begin{align}
    \frac{\sum_{j \in \mathcal{D}}(R(F_S(v_{src}), F_S(v_{dst}))^{(j)}x_i^{(j)}}{\sqrt{\sum_{j \in \mathcal{D}}(R(F_S(v_{src}), F_S(v_{dst}))^{(j)}}\sqrt{\sum_{j \in \mathcal{D}}x^{(j)}}}.
\end{align}

To assign node features, we follow the same procedure as before, where for node $v$ the associated structure feature $F_S(v)$ is used to make predictions. The generated set of features $\tilde{F}_{\mathcal{V}}$, and $\tilde{F}_{\mathcal{E}}$ are then assigned to individual nodes and edges in the generated graph structure $\tilde{S}$ using the ranking mechanism above, where ties are assigned randomly. More detail is provided in the Appendix \ref{sec:aligner}

\section{Experiments}
\label{sec:experiments}
In this section, we introduce a set of experiments to show case the effectiveness of the proposed framework for generating real-world graphs. 

\subsection{Methods}
Our method involves fitting a single large graph and learning a parametric model that can be used to generate graphs on the same scale or larger. We compare with the following baselines:

\begin{itemize}
    \item \textbf{Random}: We generate graph structures using the Erdős–Rényi model, along with a random feature generator with  ranges fitted to the original feature dimension. This model is integrated into our proposed framework.
    \item \textbf{Graphworld} \citep{palowitch2022graphworld}: is a recent method for generating arbitrary graphs using the degree corrected stochastic block model (SBM). \textbf{Note**:} \textit{we improve this method and add a fitting step that fits the model onto the underlying dataset.}
\end{itemize}

\subsection{Dataset Details}
\begin{table}[h]
\renewcommand{\arraystretch}{1.0}
\caption{\label{tab:datasetdetails}Dataset sizes used through out experiments.}
    \vspace{-4.0mm}
    \begin{center}
        \resizebox{0.5\columnwidth}{!}{
        \begin{tabular}{cccccc}
         \toprule
         ID & Dataset  & $\#$ nodes & $\#$ edges & $\#$ features\\
         \midrule
         1 & \href{https://github.com/IBM/TabFormer}{Tabformer} &  106482& 978288 & 5 &\\
         \midrule
         2 & \href{https://www.kaggle.com/c/ieee-fraud-detection}{IEEE-Fraud} & 17289 & 52008& 48& \\
         \midrule
         3 & \href{https://www.kaggle.com/datasets/ealaxi/Paysim1}{Paysim} & 9075669 &  6362620& 8&\\
         \midrule
         4 & \href{https://www.kaggle.com/datasets/kartik2112/fraud-detection}{Credit} & 1666 & 476414 & 283 &\\
         \midrule
         5 & \href{https://www.kaggle.com/datasets/ycanario/home-insurance}{Home-Credit} & 9999 & 2835954 & 16& \\
         \midrule
         6 & \href{https://www.kaggle.com/datasets/mhdzahier/travel-insurance}{Travel-Insurance} & 1986 & 172220 & 9 & \\
         \midrule
         7 & \href{https://ogb.stanford.edu/docs/lsc/mag240m/}{MAG240m} & 244160499 & 1728364232 & 768 & \\
         \midrule
         8 & \href{https://ogb.stanford.edu/docs/nodeprop/\#ogbn-mag}{OGBN-MAG} & 1939743 & 211111007 & 128 & \\
         \midrule
         9 & \href{https://relational.fit.cvut.cz/dataset/CORA}{Cora} & 2708 & 5429 & 1433 & \\
         \bottomrule
        \end{tabular}%
        }
    \end{center}
\vspace{-7.0mm}
\end{table}
%

The datasets used in the experiments are summarized in Table \ref{tab:datasetdetails}. They have different graph sizes and varying numbers of features. The steps used to construct the graphs are detailed in Table \ref{tab:data_construction} in the Appendix \ref{sec:ddetails}.

\subsection{Metrics \& Evaluation}
\vspace{-2.0mm}
Metrics that operate on a distribution of graphs such as maximum mean discrepancy (MMD)\cite{o2021evaluation} and comparisons in the latent space \cite{NEURIPS2022_3309b411} have scaling limitations, and are more suitable for comparing sets of relatively small graphs. Instead, we use a set of metrics to assess the quality of the generated graph structure and features, specifically for single graphs of varying scale, where the generated graph can be much larger than the original one:
\vspace{-2.0mm}

\begin{itemize}
    \item Degree Dist.: in networks the degree of a node is the number of connections it has with other nodes, and the degree-distribution of the network is the distribution of these degrees over the whole network. For example, a graph has a power-law if the number of nodes $N_d$ with degree $d$ is given by $N_d \propto d^{-\alpha}$ where $\alpha$ is the power law exponent. 
    \item Hop-plot: The diameter of a graph is $D$ if every pair of nodes can be connected by a path of length at most $D$ edges. As this metric is susceptible to outliers often an alternative metric called \textit{effective diameter} is used, which is the minimum number of links in which a fraction of all pairs of nodes can be reached each other. A hop-plot extends the notion of diameter by plotting reachable pairs $d(h)$ within $h$ hops.
    \item Feature Corr.: We consider the correlation between columns of the features in the dataset. For correlation between continuous columns, we use the standard Pearson correlation. Between continuous and categorical columns we consider the \textit{correlation ratio} \citep{fisher1992statistical}, and between categorical columns we consider using the Theil's U \citep{shannon1948mathematical} as a metric, which measures the conditional entropy between two variables.
    \item Degree-Feat Dist-Dist: We consider the joint degree distribution and feature distribution as a measure of graph feature+structural similarity. This metric computes the JS-divergence between the joint distribution over the generated graph and the real graph.  
\end{itemize}

\subsection{Results}
\vspace{-2.0mm}
\begin{table}
\renewcommand{\arraystretch}{1.0}
\caption{\label{tab:Comparison} Comparison across different datasets and baseline models. $\uparrow$ denotes higher is better and $\downarrow$ denotes lower is better. }
\begin{center}
\resizebox{.9\columnwidth}{!}{
\begin{tabular}{ll | ccc}
\toprule
& &  \multicolumn{3}{c}{\textbf{Metric}} \\
\textbf{Dataset} & \textbf{Method} & \textbf{Degree Dist.} $\uparrow$ & \textbf{Feature Corr.} $\uparrow$ & \textbf{Degree-Feat Dist-Dist} $\downarrow$ \\
\hline

\multirow{3}{*}{\textbf{Tabformer}}
& random & {0.8099}  & {0.3931}  & {0.8213}  \\
& graphworld & {0.2836}  & {0.3609}  & {0.8248}  \\
& ours & \textbf{{0.9904}}  & \textbf{{0.9302}}  & \textbf{{0.2620}}  \\\hline
\multirow{3}{*}{\textbf{IEEE-Fraud}}
&  random & {0.9620}  & {0.2120}  & {0.4335}  \\
& graphworld & {0.1010}  & {0.4179}  & {0.8272}  \\
& ours & \textbf{{0.9865}}  & \textbf{{0.5724}}  & \textbf{{0.2359}}   \\\hline
\multirow{3}{*}{\textbf{Credit}}
&  random & {0.0434}  & {0.8370}  & {0.6520}  \\
& graphworld & {0.3556}  & {0.8352}  & {0.7584}  \\
& ours & \textbf{{0.5178}}  & \textbf{{0.8558}}  & \textbf{{0.5516}}  \\\hline
\multirow{3}{*}{\textbf{Paysim}}
& random & {0.6711}  & {0.4833}  & {0.5155}  \\
& graphworld & {0.6547}  & {0.4115}  & {0.3453}  \\
& ours & \textbf{{0.9602}}  & \textbf{{0.7500}}  & \textbf{{0.2630}}\\
\hline

\end{tabular}
}
\end{center}
\vspace{-7.0mm}
\end{table}
In Table \ref{tab:Comparison} we summarize the comparison of the proposed framework across different datasets presented in Table \ref{tab:datasetdetails}. From the table, we can see that the synthetic data quality generated using our method outperforms the two baseline models. It is worth noting that we modified graphworld \citep{palowitch2022graphworld} to fit the underlying data. Additionally, this method can be integrated within the proposed framework where the structural generator is a SBM model, the feature generators are multi-variate Gaussian's and the aligner is a random aligner. We do not provide a comparison with methods such as \cite{you2018graphrnn, liao2019gran} as 1) these methods require multiple graphs and we are providing comparisons for single graph generation and 2) these methods do not scale beyond 1000s of nodes as previously mentioned.  
\begin{figure}[ht]
    \hspace*{-1cm} 
    \centering
    \includegraphics[scale=0.35]{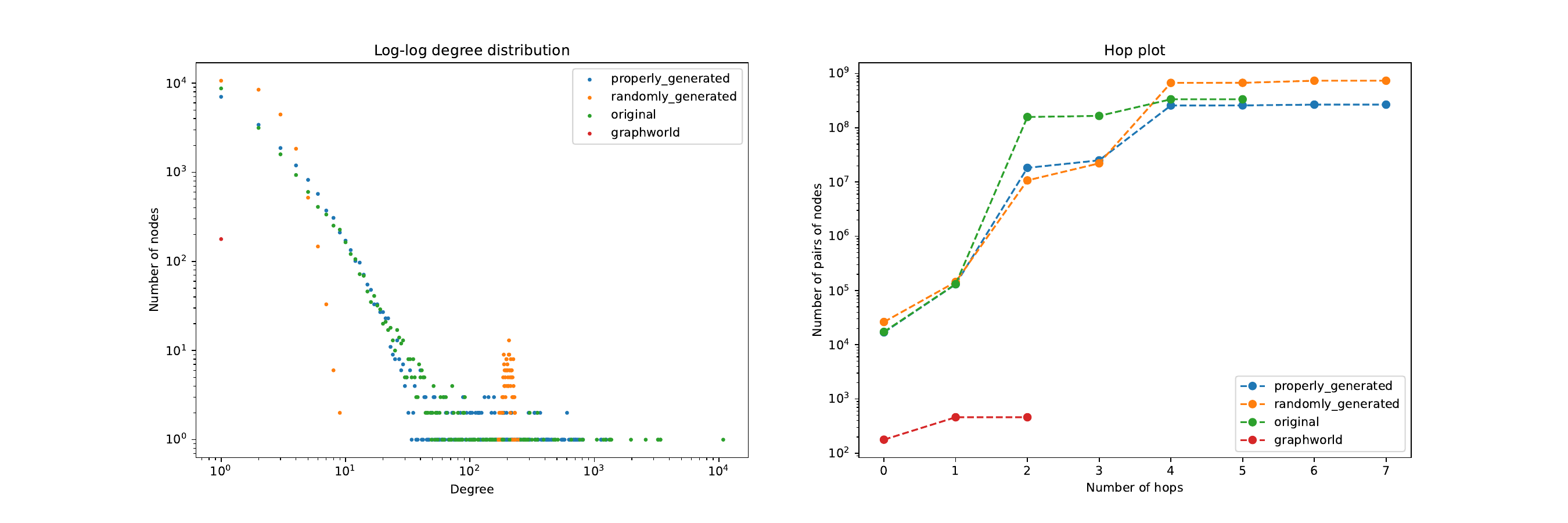}
    \vspace{-7.5mm}
    \caption{Comparison of degree distribution (left) and hop plot (right) across our proposed method (properly generated) and other baselines.}
    \label{fig:ddhop}
    \vspace{-2.0mm}
\end{figure}

We also qualitatively assess the quality of the generators compared to the baselines. From Figures \ref{fig:ddhop}, we show visualization comparing across the original graph distribution, ours, and baseline models, demonstrating the effectiveness of each component in the generation step. For example, in Figure \ref{fig:ddhop} we see that the fitted generator best resembles the long tail degree distribution of the original dataset (typically observed in social networks), whereas graphworld and randomly generated fail to capture this scaling property.

\subsection{Big Graph Generation}
\vspace{-2.0mm}

\label{sec:biggraph}
To demonstrate the scalability of the framework on generating large graphs we apply our framework to the largest publicly available dataset MAG240m \cite{wang2020microsoft}. We scaled the entire MAG240m dataset which is heterogeneous to various scales summarized in Table \ref{tab:mag_scale}. In this table the number of nodes is scaled linearly and the number of edges cubically.
All measurements were done on the same machine with 8 NVIDIA V100 16GB GPUs. We present the average results as they are highly dependent on the machine's IO speed.

\begin{table}[h]
\caption{\label{tab:mag_scale} Synthetic MAG240m \cite{wang2020microsoft} generation timings}
\vspace{-5.0mm}
\begin{center}
\resizebox{\columnwidth}{!}{

\begin{tabular}{c|cc|cc|cc|cc}
\hline
\multirow{2}{*}{scale} & \multirow{2}{*}{total nodes} & \multirow{2}{*}{total edges} & \multicolumn{2}{c}{structural part} & \multicolumn{2}{c}{tabular part} & \multicolumn{2}{c}{total} \\
                       &                              &                              & time              & memory          & time              & features     & time            & memory  \\ \hline
1x                     & 256e6                        & 1.1e9                        & $\sim$ 1 min      & $\sim$ 8G       & $\sim$ 4 min      & 134.2e6      & $\sim$ 5 min    & 203G    \\
2x                     & 536e6                        & 9e9                          & $\sim$ 5 min      & $\sim$ 64G      & $\sim$ 7 min      & 268.4e6      & $\sim$ 12 min   & 454G    \\
4x                     & 1.1e9                        & 69e9                         & $\sim$ 50 min     & $\sim$ 512G     & $\sim$ 20 min     & 536.9e6      & $\sim$ 70 min   & 1292G   \\
8x                     & 2.1e9                        & 550e9                        & $\sim$ 441 min    & $\sim$ 4096G    & $\sim$ 42 min     & 1.1e9        & $\sim$ 483 min  & 5656G   \\
10x                    & 4.3e9                        & 1.1e12                       & $\sim$ 895 min    & $\sim$ 8000G    & $\sim$ 77 min     & 2.1e9        & $\sim$ 972 min  & 9950G   \\ \hline
\end{tabular}

}
\vspace{-5.0mm}
\end{center}

\end{table}

\subsection{Additional Experiments}
\vspace{-3.0mm}

A series of additional experiments are conducted to evaluate the performance of the proposed method in the Appendix \ref{sec:additional_experiments}. In this section, a throughput analysis comparison on GNNs using the generated data is done in \ref{sec:gnn_perf}, a study on how the metrics change across different scales in \ref{sec:add_scaling_exp}, an ablation study on the framework components in \ref{sec:ablation}, a preliminary study on downstream task and pre-training for node-classification and edge-classification in \ref{sec:downstream}, a synthetic data study to determine when it is important to use both the structure and graph features, as well as when aligning them is important in \ref{sec:synth_ablation}, a comparison on structural features used for aligning in \ref{sec:aligncomparison}, a comparison of graph structural generators in \ref{sec:structgencomparison}, and a few others.

\section{Conclusion}
\vspace{-3.0mm}

    In this paper, we proposed a flexible framework for generating graph datasets that consist of (1) a structure describing how nodes are connected to other nodes and (2) tabular data associated with nodes and edges. We showed that our method is general and can be scaled to generate large-scale graph datasets. Such a method can serve many real-world use cases such as data anonymization, as a tool to benchmark GNN models by profiling them on arbitrary-sized graph datasets, as well as potentially using the generated synthetic datasets to improve the accuracy of GNN by first pre-training followed by fine-tune on the original dataset. Future directions can include investigating these directions and improving upon the framework. 

\clearpage
\bibliographystyle{ACM-Reference-Format}
\bibliography{bib}


\begin{thebibliography}{45}


\ifx \showCODEN    \undefined \def \showCODEN     #1{\unskip}     \fi
\ifx \showDOI      \undefined \def \showDOI       #1{#1}\fi
\ifx \showISBNx    \undefined \def \showISBNx     #1{\unskip}     \fi
\ifx \showISBNxiii \undefined \def \showISBNxiii  #1{\unskip}     \fi
\ifx \showISSN     \undefined \def \showISSN      #1{\unskip}     \fi
\ifx \showLCCN     \undefined \def \showLCCN      #1{\unskip}     \fi
\ifx \shownote     \undefined \def \shownote      #1{#1}          \fi
\ifx \showarticletitle \undefined \def \showarticletitle #1{#1}   \fi
\ifx \showURL      \undefined \def \showURL       {\relax}        \fi
\providecommand\bibfield[2]{#2}
\providecommand\bibinfo[2]{#2}
\providecommand\natexlab[1]{#1}
\providecommand\showeprint[2][]{arXiv:#2}

\bibitem[\protect\citeauthoryear{Abbe}{Abbe}{2017}]%
        {abbe2017community}
\bibfield{author}{\bibinfo{person}{Emmanuel Abbe}.}
  \bibinfo{year}{2017}\natexlab{}.
\newblock \showarticletitle{Community detection and stochastic block models:
  recent developments}.
\newblock \bibinfo{journal}{\emph{The Journal of Machine Learning Research}}
  \bibinfo{volume}{18}, \bibinfo{number}{1} (\bibinfo{year}{2017}),
  \bibinfo{pages}{6446--6531}.
\newblock


\bibitem[\protect\citeauthoryear{Albert and Barab\'asi}{Albert and
  Barab\'asi}{2000}]%
        {PhysRevLett.85.5234}
\bibfield{author}{\bibinfo{person}{R\'eka Albert} {and}
  \bibinfo{person}{Albert-L\'aszl\'o Barab\'asi}.}
  \bibinfo{year}{2000}\natexlab{}.
\newblock \showarticletitle{Topology of Evolving Networks: Local Events and
  Universality}.
\newblock \bibinfo{journal}{\emph{Phys. Rev. Lett.}}  \bibinfo{volume}{85}
  (\bibinfo{date}{Dec} \bibinfo{year}{2000}), \bibinfo{pages}{5234--5237}.
\newblock
Issue 24.
\urldef\tempurl%
\url{https://doi.org/10.1103/PhysRevLett.85.5234}
\showDOI{\tempurl}


\bibitem[\protect\citeauthoryear{Albert and Barab{\'a}si}{Albert and
  Barab{\'a}si}{2002}]%
        {albert2002statistical}
\bibfield{author}{\bibinfo{person}{R{\'e}ka Albert} {and}
  \bibinfo{person}{Albert-L{\'a}szl{\'o} Barab{\'a}si}.}
  \bibinfo{year}{2002}\natexlab{}.
\newblock \showarticletitle{Statistical mechanics of complex networks}.
\newblock \bibinfo{journal}{\emph{Reviews of modern physics}}
  \bibinfo{volume}{74}, \bibinfo{number}{1} (\bibinfo{year}{2002}),
  \bibinfo{pages}{47}.
\newblock


\bibitem[\protect\citeauthoryear{Bojchevski, Shchur, Zügner, and
  Günnemann}{Bojchevski et~al\mbox{.}}{2018}]%
        {netgan}
\bibfield{author}{\bibinfo{person}{Aleksandar Bojchevski},
  \bibinfo{person}{Oleksandr Shchur}, \bibinfo{person}{Daniel Zügner}, {and}
  \bibinfo{person}{Stephan Günnemann}.} \bibinfo{year}{2018}\natexlab{}.
\newblock \showarticletitle{NetGAN: Generating Graphs via Random Walks}.
\newblock  (\bibinfo{year}{2018}).
\newblock
\urldef\tempurl%
\url{https://doi.org/10.48550/ARXIV.1803.00816}
\showDOI{\tempurl}


\bibitem[\protect\citeauthoryear{Borisov, Leemann, Se{\ss}ler, Haug, Pawelczyk,
  and Kasneci}{Borisov et~al\mbox{.}}{2022}]%
        {borisov2022deep}
\bibfield{author}{\bibinfo{person}{Vadim Borisov}, \bibinfo{person}{Tobias
  Leemann}, \bibinfo{person}{Kathrin Se{\ss}ler}, \bibinfo{person}{Johannes
  Haug}, \bibinfo{person}{Martin Pawelczyk}, {and} \bibinfo{person}{Gjergji
  Kasneci}.} \bibinfo{year}{2022}\natexlab{}.
\newblock \showarticletitle{Deep neural networks and tabular data: A survey}.
\newblock \bibinfo{journal}{\emph{IEEE Transactions on Neural Networks and
  Learning Systems}} (\bibinfo{year}{2022}).
\newblock


\bibitem[\protect\citeauthoryear{Bronstein, Bruna, LeCun, Szlam, and
  Vandergheynst}{Bronstein et~al\mbox{.}}{2017}]%
        {bronstein2017geometric}
\bibfield{author}{\bibinfo{person}{Michael~M Bronstein}, \bibinfo{person}{Joan
  Bruna}, \bibinfo{person}{Yann LeCun}, \bibinfo{person}{Arthur Szlam}, {and}
  \bibinfo{person}{Pierre Vandergheynst}.} \bibinfo{year}{2017}\natexlab{}.
\newblock \showarticletitle{Geometric deep learning: going beyond euclidean
  data}.
\newblock \bibinfo{journal}{\emph{IEEE Signal Processing Magazine}}
  \bibinfo{volume}{34}, \bibinfo{number}{4} (\bibinfo{year}{2017}),
  \bibinfo{pages}{18--42}.
\newblock


\bibitem[\protect\citeauthoryear{Chakrabarti, Zhan, and Faloutsos}{Chakrabarti
  et~al\mbox{.}}{2004a}]%
        {chakrabarti2004r}
\bibfield{author}{\bibinfo{person}{Deepayan Chakrabarti},
  \bibinfo{person}{Yiping Zhan}, {and} \bibinfo{person}{Christos Faloutsos}.}
  \bibinfo{year}{2004}\natexlab{a}.
\newblock \showarticletitle{R-MAT: A recursive model for graph mining}. In
  \bibinfo{booktitle}{\emph{Proceedings of the 2004 SIAM International
  Conference on Data Mining}}. SIAM, \bibinfo{pages}{442--446}.
\newblock


\bibitem[\protect\citeauthoryear{Chakrabarti, Zhan, and Faloutsos}{Chakrabarti
  et~al\mbox{.}}{2004b}]%
        {rmat}
\bibfield{author}{\bibinfo{person}{Deepayan Chakrabarti},
  \bibinfo{person}{Yiping Zhan}, {and} \bibinfo{person}{Christos Faloutsos}.}
  \bibinfo{year}{2004}\natexlab{b}.
\newblock \showarticletitle{R-MAT: A recursive model for graph mining}.
\newblock \bibinfo{journal}{\emph{SIAM Proceedings Series}}
  \bibinfo{volume}{6}.
\newblock
\urldef\tempurl%
\url{https://doi.org/10.1137/1.9781611972740.43}
\showDOI{\tempurl}


\bibitem[\protect\citeauthoryear{Chen and Guestrin}{Chen and Guestrin}{2016}]%
        {chen2016xgboost}
\bibfield{author}{\bibinfo{person}{Tianqi Chen} {and} \bibinfo{person}{Carlos
  Guestrin}.} \bibinfo{year}{2016}\natexlab{}.
\newblock \showarticletitle{Xgboost: A scalable tree boosting system}. In
  \bibinfo{booktitle}{\emph{Proceedings of the 22nd acm sigkdd international
  conference on knowledge discovery and data mining}}.
  \bibinfo{pages}{785--794}.
\newblock


\bibitem[\protect\citeauthoryear{Erd\"{o}s and R\'{e}nyi}{Erd\"{o}s and
  R\'{e}nyi}{1959}]%
        {erdos59a}
\bibfield{author}{\bibinfo{person}{P. Erd\"{o}s} {and} \bibinfo{person}{A.
  R\'{e}nyi}.} \bibinfo{year}{1959}\natexlab{}.
\newblock \showarticletitle{On Random Graphs I}.
\newblock \bibinfo{journal}{\emph{Publicationes Mathematicae Debrecen}}
  \bibinfo{volume}{6} (\bibinfo{year}{1959}), \bibinfo{pages}{290}.
\newblock


\bibitem[\protect\citeauthoryear{Faez, Ommi, Baghshah, and Rabiee}{Faez
  et~al\mbox{.}}{2020}]%
        {dgg_survy}
\bibfield{author}{\bibinfo{person}{Faezeh Faez}, \bibinfo{person}{Yassaman
  Ommi}, \bibinfo{person}{Mahdieh~Soleymani Baghshah}, {and}
  \bibinfo{person}{Hamid~R. Rabiee}.} \bibinfo{year}{2020}\natexlab{}.
\newblock \showarticletitle{Deep Graph Generators: {A} Survey}.
\newblock \bibinfo{journal}{\emph{CoRR}}  \bibinfo{volume}{abs/2012.15544}
  (\bibinfo{year}{2020}).
\newblock
\showeprint[arXiv]{2012.15544}
\urldef\tempurl%
\url{https://arxiv.org/abs/2012.15544}
\showURL{%
\tempurl}


\bibitem[\protect\citeauthoryear{Fisher}{Fisher}{1992}]%
        {fisher1992statistical}
\bibfield{author}{\bibinfo{person}{Ronald~Aylmer Fisher}.}
  \bibinfo{year}{1992}\natexlab{}.
\newblock \showarticletitle{Statistical methods for research workers}.
\newblock In \bibinfo{booktitle}{\emph{Breakthroughs in statistics}}.
  \bibinfo{publisher}{Springer}, \bibinfo{pages}{66--70}.
\newblock


\bibitem[\protect\citeauthoryear{Goodfellow, Pouget-Abadie, Mirza, Xu,
  Warde-Farley, Ozair, Courville, and Bengio}{Goodfellow et~al\mbox{.}}{2014}]%
        {goodfellow2014generative}
\bibfield{author}{\bibinfo{person}{Ian Goodfellow}, \bibinfo{person}{Jean
  Pouget-Abadie}, \bibinfo{person}{Mehdi Mirza}, \bibinfo{person}{Bing Xu},
  \bibinfo{person}{David Warde-Farley}, \bibinfo{person}{Sherjil Ozair},
  \bibinfo{person}{Aaron Courville}, {and} \bibinfo{person}{Yoshua Bengio}.}
  \bibinfo{year}{2014}\natexlab{}.
\newblock \showarticletitle{Generative adversarial nets}.
\newblock \bibinfo{journal}{\emph{Advances in neural information processing
  systems}}  \bibinfo{volume}{27} (\bibinfo{year}{2014}).
\newblock


\bibitem[\protect\citeauthoryear{Grover and Leskovec}{Grover and
  Leskovec}{2016}]%
        {grover2016node2vec}
\bibfield{author}{\bibinfo{person}{Aditya Grover} {and} \bibinfo{person}{Jure
  Leskovec}.} \bibinfo{year}{2016}\natexlab{}.
\newblock \showarticletitle{node2vec: Scalable feature learning for networks}.
  In \bibinfo{booktitle}{\emph{Proceedings of the 22nd ACM SIGKDD international
  conference on Knowledge discovery and data mining}}.
  \bibinfo{pages}{855--864}.
\newblock


\bibitem[\protect\citeauthoryear{Grover, Zweig, and Ermon}{Grover
  et~al\mbox{.}}{2019}]%
        {grover2019graphite}
\bibfield{author}{\bibinfo{person}{Aditya Grover}, \bibinfo{person}{Aaron
  Zweig}, {and} \bibinfo{person}{Stefano Ermon}.}
  \bibinfo{year}{2019}\natexlab{}.
\newblock \showarticletitle{Graphite: Iterative generative modeling of graphs}.
  In \bibinfo{booktitle}{\emph{International conference on machine learning}}.
  PMLR, \bibinfo{pages}{2434--2444}.
\newblock


\bibitem[\protect\citeauthoryear{Hamilton, Ying, and Leskovec}{Hamilton
  et~al\mbox{.}}{2017}]%
        {hamilton2017representation}
\bibfield{author}{\bibinfo{person}{William~L Hamilton}, \bibinfo{person}{Rex
  Ying}, {and} \bibinfo{person}{Jure Leskovec}.}
  \bibinfo{year}{2017}\natexlab{}.
\newblock \showarticletitle{Representation learning on graphs: Methods and
  applications}.
\newblock \bibinfo{journal}{\emph{arXiv preprint arXiv:1709.05584}}
  (\bibinfo{year}{2017}).
\newblock


\bibitem[\protect\citeauthoryear{Hu, Fey, Ren, Nakata, Dong, and Leskovec}{Hu
  et~al\mbox{.}}{2021}]%
        {hu2021ogb}
\bibfield{author}{\bibinfo{person}{Weihua Hu}, \bibinfo{person}{Matthias Fey},
  \bibinfo{person}{Hongyu Ren}, \bibinfo{person}{Maho Nakata},
  \bibinfo{person}{Yuxiao Dong}, {and} \bibinfo{person}{Jure Leskovec}.}
  \bibinfo{year}{2021}\natexlab{}.
\newblock \showarticletitle{Ogb-lsc: A large-scale challenge for machine
  learning on graphs}.
\newblock \bibinfo{journal}{\emph{arXiv preprint arXiv:2103.09430}}
  (\bibinfo{year}{2021}).
\newblock


\bibitem[\protect\citeauthoryear{Hu, Fey, Zitnik, Dong, Ren, Liu, Catasta, and
  Leskovec}{Hu et~al\mbox{.}}{2020}]%
        {hu2020open}
\bibfield{author}{\bibinfo{person}{Weihua Hu}, \bibinfo{person}{Matthias Fey},
  \bibinfo{person}{Marinka Zitnik}, \bibinfo{person}{Yuxiao Dong},
  \bibinfo{person}{Hongyu Ren}, \bibinfo{person}{Bowen Liu},
  \bibinfo{person}{Michele Catasta}, {and} \bibinfo{person}{Jure Leskovec}.}
  \bibinfo{year}{2020}\natexlab{}.
\newblock \showarticletitle{Open graph benchmark: Datasets for machine learning
  on graphs}.
\newblock \bibinfo{journal}{\emph{Advances in neural information processing
  systems}}  \bibinfo{volume}{33} (\bibinfo{year}{2020}),
  \bibinfo{pages}{22118--22133}.
\newblock


\bibitem[\protect\citeauthoryear{Kingma and Ba}{Kingma and Ba}{2014}]%
        {kingma2014adam}
\bibfield{author}{\bibinfo{person}{Diederik~P Kingma} {and}
  \bibinfo{person}{Jimmy Ba}.} \bibinfo{year}{2014}\natexlab{}.
\newblock \showarticletitle{Adam: A method for stochastic optimization}.
\newblock \bibinfo{journal}{\emph{arXiv preprint arXiv:1412.6980}}
  (\bibinfo{year}{2014}).
\newblock


\bibitem[\protect\citeauthoryear{Kingma and Welling}{Kingma and
  Welling}{2013}]%
        {kingma2013auto}
\bibfield{author}{\bibinfo{person}{Diederik~P Kingma} {and}
  \bibinfo{person}{Max Welling}.} \bibinfo{year}{2013}\natexlab{}.
\newblock \showarticletitle{Auto-encoding variational bayes}.
\newblock \bibinfo{journal}{\emph{arXiv preprint arXiv:1312.6114}}
  (\bibinfo{year}{2013}).
\newblock


\bibitem[\protect\citeauthoryear{Kipf and Welling}{Kipf and Welling}{2016a}]%
        {kipf2016semi}
\bibfield{author}{\bibinfo{person}{Thomas~N Kipf} {and} \bibinfo{person}{Max
  Welling}.} \bibinfo{year}{2016}\natexlab{a}.
\newblock \showarticletitle{Semi-supervised classification with graph
  convolutional networks}.
\newblock \bibinfo{journal}{\emph{arXiv preprint arXiv:1609.02907}}
  (\bibinfo{year}{2016}).
\newblock


\bibitem[\protect\citeauthoryear{Kipf and Welling}{Kipf and Welling}{2016b}]%
        {kipf2016variational}
\bibfield{author}{\bibinfo{person}{Thomas~N Kipf} {and} \bibinfo{person}{Max
  Welling}.} \bibinfo{year}{2016}\natexlab{b}.
\newblock \showarticletitle{Variational graph auto-encoders}.
\newblock \bibinfo{journal}{\emph{arXiv preprint arXiv:1611.07308}}
  (\bibinfo{year}{2016}).
\newblock


\bibitem[\protect\citeauthoryear{Leskovec, Chakrabarti, Kleinberg, and
  Faloutsos}{Leskovec et~al\mbox{.}}{2005}]%
        {10.1007/11564126_17}
\bibfield{author}{\bibinfo{person}{Jurij Leskovec}, \bibinfo{person}{Deepayan
  Chakrabarti}, \bibinfo{person}{Jon Kleinberg}, {and}
  \bibinfo{person}{Christos Faloutsos}.} \bibinfo{year}{2005}\natexlab{}.
\newblock \showarticletitle{Realistic, Mathematically Tractable Graph
  Generation and Evolution, Using Kronecker Multiplication}. In
  \bibinfo{booktitle}{\emph{Knowledge Discovery in Databases: PKDD 2005}},
  \bibfield{editor}{\bibinfo{person}{Al{\'i}pio~M{\'a}rio Jorge},
  \bibinfo{person}{Lu{\'i}s Torgo}, \bibinfo{person}{Pavel Brazdil},
  \bibinfo{person}{Rui Camacho}, {and} \bibinfo{person}{Jo{\~a}o Gama}} (Eds.).
  \bibinfo{publisher}{Springer Berlin Heidelberg}, \bibinfo{address}{Berlin,
  Heidelberg}, \bibinfo{pages}{133--145}.
\newblock
\showISBNx{978-3-540-31665-7}


\bibitem[\protect\citeauthoryear{Leskovec, Chakrabarti, Kleinberg, Faloutsos,
  and Ghahramani}{Leskovec et~al\mbox{.}}{2010}]%
        {leskovec2010kronecker}
\bibfield{author}{\bibinfo{person}{Jure Leskovec}, \bibinfo{person}{Deepayan
  Chakrabarti}, \bibinfo{person}{Jon Kleinberg}, \bibinfo{person}{Christos
  Faloutsos}, {and} \bibinfo{person}{Zoubin Ghahramani}.}
  \bibinfo{year}{2010}\natexlab{}.
\newblock \showarticletitle{Kronecker graphs: an approach to modeling
  networks.}
\newblock \bibinfo{journal}{\emph{Journal of Machine Learning Research}}
  \bibinfo{volume}{11}, \bibinfo{number}{2} (\bibinfo{year}{2010}).
\newblock


\bibitem[\protect\citeauthoryear{Liao, Li, Song, Wang, Nash, Hamilton,
  Duvenaud, Urtasun, and Zemel}{Liao et~al\mbox{.}}{2019}]%
        {liao2019gran}
\bibfield{author}{\bibinfo{person}{Renjie Liao}, \bibinfo{person}{Yujia Li},
  \bibinfo{person}{Yang Song}, \bibinfo{person}{Shenlong Wang},
  \bibinfo{person}{Charlie Nash}, \bibinfo{person}{William~L. Hamilton},
  \bibinfo{person}{David Duvenaud}, \bibinfo{person}{Raquel Urtasun}, {and}
  \bibinfo{person}{Richard Zemel}.} \bibinfo{year}{2019}\natexlab{}.
\newblock \showarticletitle{Efficient Graph Generation with Graph Recurrent
  Attention Networks}. In \bibinfo{booktitle}{\emph{NeurIPS}}.
\newblock


\bibitem[\protect\citeauthoryear{Liben-Nowell and Kleinberg}{Liben-Nowell and
  Kleinberg}{2007}]%
        {liben2007link}
\bibfield{author}{\bibinfo{person}{David Liben-Nowell} {and}
  \bibinfo{person}{Jon Kleinberg}.} \bibinfo{year}{2007}\natexlab{}.
\newblock \showarticletitle{The link-prediction problem for social networks}.
\newblock \bibinfo{journal}{\emph{Journal of the American society for
  information science and technology}} \bibinfo{volume}{58},
  \bibinfo{number}{7} (\bibinfo{year}{2007}), \bibinfo{pages}{1019--1031}.
\newblock


\bibitem[\protect\citeauthoryear{Ma'ayan, Lipshtat, Iyengar, and
  Sontag}{Ma'ayan et~al\mbox{.}}{2008}]%
        {ma2008proximity}
\bibfield{author}{\bibinfo{person}{Avi Ma'ayan}, \bibinfo{person}{Azi
  Lipshtat}, \bibinfo{person}{Ravi Iyengar}, {and} \bibinfo{person}{Eduardo~D
  Sontag}.} \bibinfo{year}{2008}\natexlab{}.
\newblock \showarticletitle{Proximity of intracellular regulatory networks to
  monotone systems}.
\newblock \bibinfo{journal}{\emph{IET Systems Biology}} \bibinfo{volume}{2},
  \bibinfo{number}{3} (\bibinfo{year}{2008}), \bibinfo{pages}{103--112}.
\newblock


\bibitem[\protect\citeauthoryear{Martinkus, Loukas, Perraudin, and
  Wattenhofer}{Martinkus et~al\mbox{.}}{2022}]%
        {martinkus2022spectre}
\bibfield{author}{\bibinfo{person}{Karolis Martinkus}, \bibinfo{person}{Andreas
  Loukas}, \bibinfo{person}{Nathana{\"e}l Perraudin}, {and}
  \bibinfo{person}{Roger Wattenhofer}.} \bibinfo{year}{2022}\natexlab{}.
\newblock \showarticletitle{SPECTRE: Spectral Conditioning Helps to Overcome
  the Expressivity Limits of One-shot Graph Generators}.
\newblock \bibinfo{journal}{\emph{arXiv preprint arXiv:2204.01613}}
  (\bibinfo{year}{2022}).
\newblock


\bibitem[\protect\citeauthoryear{O'Bray, Horn, Rieck, and Borgwardt}{O'Bray
  et~al\mbox{.}}{2021}]%
        {o2021evaluation}
\bibfield{author}{\bibinfo{person}{Leslie O'Bray}, \bibinfo{person}{Max Horn},
  \bibinfo{person}{Bastian Rieck}, {and} \bibinfo{person}{Karsten Borgwardt}.}
  \bibinfo{year}{2021}\natexlab{}.
\newblock \showarticletitle{Evaluation metrics for graph generative models:
  Problems, pitfalls, and practical solutions}.
\newblock \bibinfo{journal}{\emph{arXiv preprint arXiv:2106.01098}}
  (\bibinfo{year}{2021}).
\newblock


\bibitem[\protect\citeauthoryear{Palowitch, Tsitsulin, Mayer, and
  Perozzi}{Palowitch et~al\mbox{.}}{2022}]%
        {palowitch2022graphworld}
\bibfield{author}{\bibinfo{person}{John Palowitch}, \bibinfo{person}{Anton
  Tsitsulin}, \bibinfo{person}{Brandon Mayer}, {and} \bibinfo{person}{Bryan
  Perozzi}.} \bibinfo{year}{2022}\natexlab{}.
\newblock \showarticletitle{GraphWorld: Fake Graphs Bring Real Insights for
  GNNs}.
\newblock \bibinfo{journal}{\emph{arXiv preprint arXiv:2203.00112}}
  (\bibinfo{year}{2022}).
\newblock


\bibitem[\protect\citeauthoryear{Park and Kim}{Park and Kim}{2017}]%
        {park2017trilliong}
\bibfield{author}{\bibinfo{person}{Himchan Park} {and} \bibinfo{person}{Min-Soo
  Kim}.} \bibinfo{year}{2017}\natexlab{}.
\newblock \showarticletitle{TrillionG: A trillion-scale synthetic graph
  generator using a recursive vector model}. In
  \bibinfo{booktitle}{\emph{Proceedings of the 2017 ACM International
  Conference on Management of Data}}. ACM, \bibinfo{pages}{913--928}.
\newblock


\bibitem[\protect\citeauthoryear{Parzen}{Parzen}{1962}]%
        {parzen1962estimation}
\bibfield{author}{\bibinfo{person}{Emanuel Parzen}.}
  \bibinfo{year}{1962}\natexlab{}.
\newblock \showarticletitle{On estimation of a probability density function and
  mode}.
\newblock \bibinfo{journal}{\emph{The annals of mathematical statistics}}
  \bibinfo{volume}{33}, \bibinfo{number}{3} (\bibinfo{year}{1962}),
  \bibinfo{pages}{1065--1076}.
\newblock


\bibitem[\protect\citeauthoryear{Qiu, Dong, Ma, Li, Wang, Wang, and Tang}{Qiu
  et~al\mbox{.}}{2019}]%
        {qiu2019netsmf}
\bibfield{author}{\bibinfo{person}{Jiezhong Qiu}, \bibinfo{person}{Yuxiao
  Dong}, \bibinfo{person}{Hao Ma}, \bibinfo{person}{Jian Li},
  \bibinfo{person}{Chi Wang}, \bibinfo{person}{Kuansan Wang}, {and}
  \bibinfo{person}{Jie Tang}.} \bibinfo{year}{2019}\natexlab{}.
\newblock \showarticletitle{Netsmf: Large-scale network embedding as sparse
  matrix factorization}. In \bibinfo{booktitle}{\emph{The World Wide Web
  Conference}}. \bibinfo{pages}{1509--1520}.
\newblock


\bibitem[\protect\citeauthoryear{Rosenblatt}{Rosenblatt}{1956}]%
        {rosenblatt1956remarks}
\bibfield{author}{\bibinfo{person}{Murray Rosenblatt}.}
  \bibinfo{year}{1956}\natexlab{}.
\newblock \showarticletitle{Remarks on some nonparametric estimates of a
  density function}.
\newblock \bibinfo{journal}{\emph{The annals of mathematical statistics}}
  (\bibinfo{year}{1956}), \bibinfo{pages}{832--837}.
\newblock


\bibitem[\protect\citeauthoryear{Schlichtkrull, Kipf, Bloem, Berg, Titov, and
  Welling}{Schlichtkrull et~al\mbox{.}}{2018}]%
        {schlichtkrull2018modeling}
\bibfield{author}{\bibinfo{person}{Michael Schlichtkrull},
  \bibinfo{person}{Thomas~N Kipf}, \bibinfo{person}{Peter Bloem},
  \bibinfo{person}{Rianne van~den Berg}, \bibinfo{person}{Ivan Titov}, {and}
  \bibinfo{person}{Max Welling}.} \bibinfo{year}{2018}\natexlab{}.
\newblock \showarticletitle{Modeling relational data with graph convolutional
  networks}. In \bibinfo{booktitle}{\emph{European semantic web conference}}.
  Springer, \bibinfo{pages}{593--607}.
\newblock


\bibitem[\protect\citeauthoryear{Sen, Namata, Bilgic, Getoor, Galligher, and
  Eliassi-Rad}{Sen et~al\mbox{.}}{2008}]%
        {sen2008collective}
\bibfield{author}{\bibinfo{person}{Prithviraj Sen}, \bibinfo{person}{Galileo
  Namata}, \bibinfo{person}{Mustafa Bilgic}, \bibinfo{person}{Lise Getoor},
  \bibinfo{person}{Brian Galligher}, {and} \bibinfo{person}{Tina Eliassi-Rad}.}
  \bibinfo{year}{2008}\natexlab{}.
\newblock \showarticletitle{Collective classification in network data}.
\newblock \bibinfo{journal}{\emph{AI magazine}} \bibinfo{volume}{29},
  \bibinfo{number}{3} (\bibinfo{year}{2008}), \bibinfo{pages}{93--93}.
\newblock


\bibitem[\protect\citeauthoryear{Seshadhri, Pinar, and Kolda}{Seshadhri
  et~al\mbox{.}}{2011}]%
        {skg_hitch}
\bibfield{author}{\bibinfo{person}{C. Seshadhri}, \bibinfo{person}{Ali Pinar},
  {and} \bibinfo{person}{Tamara Kolda}.} \bibinfo{year}{2011}\natexlab{}.
\newblock \showarticletitle{A Hitchhiker's Guide to Choosing Parameters of
  Stochastic Kronecker Graphs}.
\newblock \bibinfo{journal}{\emph{CoRR}}  \bibinfo{volume}{abs/1102.5046}
  (\bibinfo{date}{01} \bibinfo{year}{2011}).
\newblock


\bibitem[\protect\citeauthoryear{Shannon}{Shannon}{1948}]%
        {shannon1948mathematical}
\bibfield{author}{\bibinfo{person}{Claude~E Shannon}.}
  \bibinfo{year}{1948}\natexlab{}.
\newblock \showarticletitle{A mathematical theory of communication, Bell
  Systems Technol}.
\newblock \bibinfo{journal}{\emph{J}} \bibinfo{volume}{27}, \bibinfo{number}{3}
  (\bibinfo{year}{1948}), \bibinfo{pages}{379--423}.
\newblock


\bibitem[\protect\citeauthoryear{Shirzad, Hassani, and Sutherland}{Shirzad
  et~al\mbox{.}}{2022}]%
        {NEURIPS2022_3309b411}
\bibfield{author}{\bibinfo{person}{Hamed Shirzad}, \bibinfo{person}{Kaveh
  Hassani}, {and} \bibinfo{person}{Danica~J. Sutherland}.}
  \bibinfo{year}{2022}\natexlab{}.
\newblock \showarticletitle{Evaluating Graph Generative Models with
  Contrastively Learned Features}. In \bibinfo{booktitle}{\emph{Advances in
  Neural Information Processing Systems}},
  \bibfield{editor}{\bibinfo{person}{S.~Koyejo}, \bibinfo{person}{S.~Mohamed},
  \bibinfo{person}{A.~Agarwal}, \bibinfo{person}{D.~Belgrave},
  \bibinfo{person}{K.~Cho}, {and} \bibinfo{person}{A.~Oh}} (Eds.),
  Vol.~\bibinfo{volume}{35}. \bibinfo{publisher}{Curran Associates, Inc.},
  \bibinfo{pages}{7783--7795}.
\newblock
\urldef\tempurl%
\url{https://proceedings.neurips.cc/paper_files/paper/2022/file/3309b4112c9f04a993f2bbdd0274bba1-Paper-Conference.pdf}
\showURL{%
\tempurl}


\bibitem[\protect\citeauthoryear{Veli{\v{c}}kovi{\'c}, Cucurull, Casanova,
  Romero, Lio, and Bengio}{Veli{\v{c}}kovi{\'c} et~al\mbox{.}}{2017}]%
        {velivckovic2017graph}
\bibfield{author}{\bibinfo{person}{Petar Veli{\v{c}}kovi{\'c}},
  \bibinfo{person}{Guillem Cucurull}, \bibinfo{person}{Arantxa Casanova},
  \bibinfo{person}{Adriana Romero}, \bibinfo{person}{Pietro Lio}, {and}
  \bibinfo{person}{Yoshua Bengio}.} \bibinfo{year}{2017}\natexlab{}.
\newblock \showarticletitle{Graph attention networks}.
\newblock \bibinfo{journal}{\emph{arXiv preprint arXiv:1710.10903}}
  (\bibinfo{year}{2017}).
\newblock


\bibitem[\protect\citeauthoryear{Wang, Wang, Huang, Song, and Li}{Wang
  et~al\mbox{.}}{2021}]%
        {wang2021fastsgg}
\bibfield{author}{\bibinfo{person}{Chaokun Wang}, \bibinfo{person}{Binbin
  Wang}, \bibinfo{person}{Bingyang Huang}, \bibinfo{person}{Shaoxu Song}, {and}
  \bibinfo{person}{Zai Li}.} \bibinfo{year}{2021}\natexlab{}.
\newblock \showarticletitle{Fastsgg: Efficient social graph generation using a
  degree distribution generation model}. In \bibinfo{booktitle}{\emph{2021 IEEE
  37th International Conference on Data Engineering (ICDE)}}. IEEE,
  \bibinfo{pages}{564--575}.
\newblock


\bibitem[\protect\citeauthoryear{Wang, Shen, Huang, Wu, Dong, and Kanakia}{Wang
  et~al\mbox{.}}{2020}]%
        {wang2020microsoft}
\bibfield{author}{\bibinfo{person}{Kuansan Wang}, \bibinfo{person}{Zhihong
  Shen}, \bibinfo{person}{Chiyuan Huang}, \bibinfo{person}{Chieh-Han Wu},
  \bibinfo{person}{Yuxiao Dong}, {and} \bibinfo{person}{Anshul Kanakia}.}
  \bibinfo{year}{2020}\natexlab{}.
\newblock \showarticletitle{Microsoft academic graph: When experts are not
  enough}.
\newblock \bibinfo{journal}{\emph{Quantitative Science Studies}}
  \bibinfo{volume}{1}, \bibinfo{number}{1} (\bibinfo{year}{2020}),
  \bibinfo{pages}{396--413}.
\newblock


\bibitem[\protect\citeauthoryear{Wills and Meyer}{Wills and Meyer}{2020}]%
        {wills2020metrics}
\bibfield{author}{\bibinfo{person}{Peter Wills} {and}
  \bibinfo{person}{Fran{\c{c}}ois~G Meyer}.} \bibinfo{year}{2020}\natexlab{}.
\newblock \showarticletitle{Metrics for graph comparison: a practitioner’s
  guide}.
\newblock \bibinfo{journal}{\emph{Plos one}} \bibinfo{volume}{15},
  \bibinfo{number}{2} (\bibinfo{year}{2020}), \bibinfo{pages}{e0228728}.
\newblock


\bibitem[\protect\citeauthoryear{Xu, Skoularidou, Cuesta-Infante, and
  Veeramachaneni}{Xu et~al\mbox{.}}{2019}]%
        {NEURIPS2019_254ed7d2}
\bibfield{author}{\bibinfo{person}{Lei Xu}, \bibinfo{person}{Maria
  Skoularidou}, \bibinfo{person}{Alfredo Cuesta-Infante}, {and}
  \bibinfo{person}{Kalyan Veeramachaneni}.} \bibinfo{year}{2019}\natexlab{}.
\newblock \showarticletitle{Modeling Tabular data using Conditional GAN}. In
  \bibinfo{booktitle}{\emph{Advances in Neural Information Processing
  Systems}}, \bibfield{editor}{\bibinfo{person}{H.~Wallach},
  \bibinfo{person}{H.~Larochelle}, \bibinfo{person}{A.~Beygelzimer},
  \bibinfo{person}{F.~d\textquotesingle Alch\'{e}-Buc},
  \bibinfo{person}{E.~Fox}, {and} \bibinfo{person}{R.~Garnett}} (Eds.),
  Vol.~\bibinfo{volume}{32}. \bibinfo{publisher}{Curran Associates, Inc.}
\newblock
\urldef\tempurl%
\url{https://proceedings.neurips.cc/paper/2019/file/254ed7d2de3b23ab10936522dd547b78-Paper.pdf}
\showURL{%
\tempurl}


\bibitem[\protect\citeauthoryear{You, Ying, Ren, Hamilton, and Leskovec}{You
  et~al\mbox{.}}{2018}]%
        {you2018graphrnn}
\bibfield{author}{\bibinfo{person}{Jiaxuan You}, \bibinfo{person}{Rex Ying},
  \bibinfo{person}{Xiang Ren}, \bibinfo{person}{William Hamilton}, {and}
  \bibinfo{person}{Jure Leskovec}.} \bibinfo{year}{2018}\natexlab{}.
\newblock \showarticletitle{Graphrnn: Generating realistic graphs with deep
  auto-regressive models}. In \bibinfo{booktitle}{\emph{International
  conference on machine learning}}. PMLR, \bibinfo{pages}{5708--5717}.
\newblock


\end{thebibliography}

\section{Broader Impact \& Limitation}
\label{sec:broaderimpact}

Graph datasets and analytics are becoming more and more pervasive among applications. With the increasing success of deep-learning in other domains (especially in image and language domains), these have not been fully extended in the graph domain. A limited number of works can generate graphs on a relatively large scale. This paper takes a step towards providing such a framework for generating large graphs with both edge and node features. This method can then be leveraged to analyze performance or share synthetic data for graph neural network models for domains where public data is limited. The proposed framework decouples the feature generation and structure generation, which may not be suitable for graphs with physical properties such as molecules. This is because the structure and features of these graphs are tightly coupled, and they must be used jointly to capture the underlying generating process. Further, we do not address model performance improvement using our generation scheme leaving this for future study.

\clearpage

\section{Aligner Additional Details}

\begin{figure}
    \centering
    \includegraphics[scale=1.05]{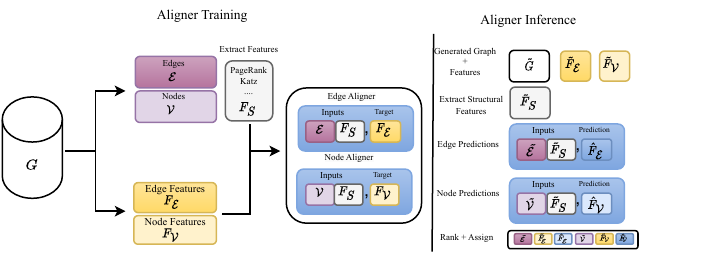}
    \caption{A system level view of how the aligner functions during training and generation (inference) to supplement the text in section \ref{sec:aligner}.}
    \label{fig:aligner}
\end{figure}

As depicted in Figure \ref{fig:aligner}, during training the aligner is trained on the input graph dataset, where the corresponding graph structure is used to obtain a series of graph related features such as PageRank, Katz Centrality, and Degree are extracted. These features and the features contained in the graph are used as a dataset to train a predictor, in which case we use an $\textrm{XGBoost}$ model. A separate model is trained for each feature and each edge, or node type. During generation, the aligner takes as input the same set of features extracted during training from the generated graph and predicts the features for each node and edge in the graph. These are then used to compute a similarity measure between the predicted feature and the generated features, from which they are then ranked and assigned accordingly to each node/edge in the graph.

\section{Additional Experiments \& Details}
\label{sec:additional_experiments}
\subsection{GNN Performance Analysis}
\label{sec:gnn_perf}
\begin{table}[h]
\renewcommand{\arraystretch}{1.0}
\caption{\label{tab:perf_comparison} Comparison of throughput for GCN \citep{kipf2016semi} (left) and GAT\citep{velivckovic2017graph} (right)  across different datasets. $\uparrow$ denotes higher is better.}
\vspace{-2.0mm}
\begin{minipage}{.5\textwidth}
\begin{center}
\resizebox{\columnwidth}{!}{
\small
\begin{tabular}{ll | cc}
\toprule
& &  \multicolumn{2}{c}{\textbf{Metric}} \\
\textbf{Dataset} & \textbf{Method} & \textbf{Rel. Timing} $\uparrow$ & \textbf{Timing}\\
\hline
\multirow{3}{*}{\textbf{Tabformer}}
& original & {1.0}  & {107.5836} \\
& random & {0.8611 $\pm 0.8611$}  & {93.3228}  \\
& ours & \textbf{{0.9376 $\pm 0.0094$}}  & \textbf{{100.1258}} \\\hline
\multirow{3}{*}{\textbf{IEEE-Fraud}}
& original & {1.0}  & {1.9833} \\
&  random & {0.7922} $\pm {0.1484}$  & {2.06162}  \\
& ours & \textbf{{0.8039 $\pm {0.1450}$}}  & \textbf{{2.04135}}   \\\hline
\multirow{3}{*}{\textbf{Credit}}
& original & {1.0}  & {28.6372} \\
&  random & {0.9668 $\pm 0.0003$}  & {26.9836}  \\
& ours & \textbf{{0.9822}$\pm 0.0083$}  & \textbf{{28.1271}} \\\hline
\multirow{3}{*}{\textbf{Paysim}}
& original & {1.0}  & {153.4566} \\
& random & {0.9302 $\pm {0.0025}$}  & {143.1298} \\
& ours & \textbf{{0.9581} $\pm {0.0020}$}  & \textbf{{159.1917}} \\
\hline
\end{tabular}

}
\end{center}
\end{minipage}
\begin{minipage}{.5\linewidth}
\begin{center}
\resizebox{\columnwidth}{!}{
\small
\begin{tabular}{ll | cc}
\toprule
& &  \multicolumn{2}{c}{\textbf{Metric}} \\
\textbf{Dataset} & \textbf{Method} & \textbf{Rel. Timing} $\uparrow$ & \textbf{Timing}\\
\hline
\multirow{3}{*}{\textbf{Tabformer}}
& original & {1.0}  & {191.6026} \\
& random & {0.6903 $\pm .0004$}  & {132.2720}  \\
& ours & \textbf{{0.8710 $\pm 0.0038$}}  & \textbf{{166.8899}} \\\hline
\multirow{3}{*}{\textbf{IEEE-Fraud}}
& original & {1.0}  & {3.9867} \\
&  random & {0.6975} $\pm {0.0795}$  & {3.41476}  \\
& ours & \textbf{{0.8003 $\pm {0.0940}$}}  & \textbf{{3.5655}}   \\\hline
\multirow{3}{*}{\textbf{Credit}}
& original & {1.0}  & {70.2702} \\
&  random & {0.9370 $\pm 0.0011$}  & {65.8476}  \\
& ours & \textbf{{0.9609}$\pm 0.0034$}  & \textbf{{67.8039}} \\\hline
\multirow{3}{*}{\textbf{Paysim}}
& original & {1.0}  & {235.1643} \\
& random & {0.9248 $\pm {0.0016}$}  & {217.4626} \\
& ours & \textbf{{0.9657} $\pm {0.0046}$}  & \textbf{{243.2182}} \\
\hline
\end{tabular}
}
\end{center}
\end{minipage}
\vspace{-3.0mm}
\end{table}

In this section, we run additional experiments analyzing GNN's throughput using our generated and randomly generated datasets. We time the throughput of the GNN network by sampling subgraphs on the original graph using a Multi-Layer Neighborhood Sampler \footnote{\href{https://docs.dgl.ai/en/0.6.x/api/python/dgl.dataloading.html\#dgl.dataloading.neighbor.MultiLayerNeighborSampler}{dgl.dataloading.neighbor.MultiLayerNeighborSampler}} and measuring the time of every epoch. The results are summarized in Table. \ref{tab:perf_comparison}. As can be seen from the results, generally, for datasets where there's a higher discrepancy in the metrics reported in Table \ref{tab:Comparison} there is a larger gap in the timing between random vs original and ours. The relative timing (Rel. Timing) is calculated by subtracting the epoch times on the original dataset from the generated and normalizing, i.e. (Rel. Timing = $1.0 - \frac{|t_{generated} - t_{original}|}{t_{original}}$

\subsection{Scaling Experiments}
\label{sec:add_scaling_exp}

In this section, we take the datasets listed in Table \ref{tab:Comparison}, as well as three other datasets and scale them to sizes [1, 2, 4, 8] evaluating them using the same metrics. The results are listed in Table \ref{tab:scale_comparison}, for certain datasets as they are scaled the metrics remain mostly the same whereas for some the metrics degrade as they are scaled which may indicate that a more expressive generator is necessary.

\begin{table}
\renewcommand{\arraystretch}{1.0}
\caption{\label{tab:scale_comparison} Comparison across different scales of datasets. $\uparrow$ denotes higher is better and $\downarrow$ denotes lower is better. In order to maintain the graph sparsity we scale nodes linearly and edges quadratically.}
\begin{center}
\resizebox{.8\columnwidth}{!}{

\begin{tabular}{cc|ccc} 
\toprule
& &  \multicolumn{3}{c}{\textbf{Metric}} \\ 
\textbf{Dataset} & \textbf{Scale} & \textbf{Degree Dist.} $\uparrow$ & \textbf{Feature Corr.} $\uparrow$ & \textbf{Degree-Feat Dist-Dist} $\downarrow$ \\ \hline 
\multirow{4}{*}{\textbf{Tabformer}}
& 8 & 0.9771 & 0.9158 & 0.5712 \\ 
& 4 & 0.9763 & 0.9281 & 0.3362 \\ 
& 2 & 0.9766 & 0.8980 & 0.3356 \\ 
\hline 
\multirow{4}{*}{\textbf{IEEE-Fraud}}
& 8 & 0.9085 & 0.6051 & 0.7714 \\ 
& 4 & 0.9846 & 0.4883 & 0.7038 \\ 
& 2 & 0.9771 & 0.6059 & 0.7208 \\ 
\hline 
\multirow{3}{*}{\textbf{Paysim}}
& 8 & 0.7088 & 0.9767 & 0.8236 \\ 
& 4 & 0.7002 & 0.9936 & 0.8180 \\ 
& 2 & 0.7516 & 0.9927 & 0.8157 \\ 
\hline 
\multirow{4}{*}{\textbf{Home Credit}}
& 8 & 0.4079 & 0.9143 & 0.8149 \\ 
& 4 & 0.5558 & 0.9271 & 0.8148 \\ 
& 2 & 0.5780 & 0.9233 & 0.8205 \\ 
& 1 & 0.5510 & 0.9205 & 0.8284 \\ 
\hline 
\multirow{4}{*}{\textbf{Travel Insurance}}
& 8 & 0.4479 & 0.8465 & 0.8164 \\ 
& 4 & 0.4902 & 0.8630 & 0.7860 \\ 
& 2 & 0.5861 & 0.8391 & 0.6892 \\ 
& 1 & 0.9209 & 0.8476 & 0.6541 \\ 
\hline 
\multirow{4}{*}{\textbf{OGBN MAG}}
& 8 & 0.9298 & 0.8334 & 0.2065 \\ 
& 4 & 0.9332 & 0.9076 & 0.2054 \\ 
& 2 & 0.9316 & 0.9080 & 0.2053 \\ 
& 1 & 0.9310 & 0.9081 & 0.2053 \\ 
\hline 
\end{tabular}

}
\end{center}
\vspace{-7.0mm}
\end{table}

\subsection{Ablation Study}
\label{sec:ablation}
We conduct an ablation study by varying the components in our proposed framework. Specifically, we substitute the feature generator with one of \{GAN, Kernel Density Estimator (KDE), Random\}, the structural generator with one of \{Ours, TrillionG \citep{park2017trilliong}, Random\} and the aligner with either \{XGBoost, Random\}   across our proposed components and evaluate the synthetic data quality. The results are presented in \ref{tab:ablation}. This showcases the benefit of providing a fitting mechanism for each component.

\begin{table*}[h]
\renewcommand{\arraystretch}{1.0}
\caption{\label{tab:ablation} Ablation study on IEEE Dataset}
\begin{center}
\resizebox{\columnwidth}{!}{
\begin{tabular}{l|l|l | c|c|c}
\toprule
& & & \multicolumn{3}{c}{\textbf{Metric}} \\
\textbf{Struct. Generator} & \textbf{Feature Generator} & \textbf{Aligner} & \textbf{Degree Dist.} $\uparrow$ & \textbf{Feature Corr.} $\uparrow$ & \textbf{Degree-Feat Dist-Dist} $\downarrow$ \\
\hline
\multirow{3}{*}{\textbf{Ours}}
& \multirow{2}{*}{\textbf{GAN}} 
& xgboost   & 0.989123 $(\pm 0.0014)$ & 0.566755 $(\pm 0.0121)$ & 0.319410\\ \cline{3-6}
& & random  & 0.989123 $(\pm 0.0014)$ & 0.566755 $(\pm 0.0121)$ & 0.330570\\ \cline{2-6}
& \multirow{2}{*}{\textbf{KDE}}
& xgboost   & 0.989123 $(\pm 0.0014)$ & 0.810647 $(\pm 0.0277)$ & 0.198234\\ \cline{3-6}
& & random  & 0.989123 $(\pm 0.0014)$ & 0.810647 $(\pm 0.0277)$ & 0.512846\\ \cline{2-6}
& \multirow{2}{*}{\textbf{Random}}
& xgboost   & 0.989123 $(\pm 0.0014)$ & 0.220692 $(\pm 0.0090)$ & 0.355433 \\ \cline{3-6}
& & random  & 0.989123 $(\pm 0.0014)$ & 0.220692 $(\pm 0.0090)$ & 0.364814\\ \hline
\multirow{3}{*}{\textbf{TrillionG \cite{park2017trilliong}}}
& \multirow{2}{*}{\textbf{GAN}} 
& xgboost   & 0.848164 $(\pm 0.0012)$ & 0.566755 $(\pm 0.0121)$ & 0.316364 \\ \cline{3-6}
& & random  & 0.848164 $(\pm 0.0012)$ & 0.566755 $(\pm 0.0121)$ & 0.423613\\ \cline{2-6}
& \multirow{2}{*}{\textbf{KDE}}
& xgboost   & 0.848164 $(\pm 0.0012)$ & 0.810647 $(\pm 0.0277)$ & 0.261013\\ \cline{3-6}
& & random  & 0.848164 $(\pm 0.0012)$ & 0.810647 $(\pm 0.0277)$ & 0.360605 \\ \cline{2-6}
& \multirow{2}{*}{\textbf{Random}}
& xgboost   & 0.848164 $(\pm 0.0012)$ & 0.220692 $(\pm 0.0090)$ & 0.434184\\ \cline{3-6}
& & random  & 0.848164 $(\pm 0.0012)$ & 0.220692 $(\pm 0.0090)$ &  0.484187\\ \hline
\multirow{3}{*}{\textbf{Random} \citep{erdos59a}}
& \multirow{2}{*}{\textbf{GAN}} 
& xgboost   & 0.962508 $(\pm 0.0011)$ & 0.566755 $(\pm 0.0121)$  & 0.420893\\ \cline{3-6}
& & random  & 0.962508 $(\pm 0.0011)$ & 0.566755 $(\pm 0.0121)$  & 0.480556\\ \cline{2-6}
& \multirow{2}{*}{\textbf{KDE}}
& xgboost   & 0.962508 $(\pm 0.0011)$ & 0.810647 $(\pm 0.0277)$ & 0.38334\\ \cline{3-6}
& & random  & 0.962508 $(\pm 0.0011)$ & 0.810647 $(\pm 0.0277)$ & 0.400145\\ \cline{2-6}
& \multirow{2}{*}{\textbf{Random}}
& xgboost   & 0.962508 $(\pm 0.0011)$ & 0.220692 $(\pm 0.009081)$ & 0.434184\\ \cline{3-6}
& & random  & 0.962508 $(\pm 0.0011)$ & 0.220692 $(\pm 0.0090)$ & 0.613355\\ 
\hline
\end{tabular}
}
\end{center}
\end{table*}

\subsection{Downstream Tasks \& Pretraining}
\label{sec:downstream}

Common downstream tasks in graphs datasets are node classification and edge classification. The proposed framework can generate both node-level and edge-level features, hence it supports these various tasks using the proposed generator. To this end, we train our synthetic data generator on both node classification and edge classification datasets, subsequently, we generate a graph of the same size for pre-training. Finally, we fine-tune on the original real dataset. The results are presented in Table \ref{tab:nc_ec}, where \textit{no-pretraining} refers to the case where we simply train on the downstream dataset.

For the node-classification task, we use the Cora \citep{sen2008collective} as a toy example which is a citation network dataset, with node labels as topics and features as multi-hot vectors; and for edge-classification task the IEEE-Fraud dataset, which contains edge features as well as labels denoting whether a particular transaction is fraudulent. For all datasets, we train the models for a maximum of 200 epochs (fine-tuning epochs + pre-training epochs) and use Adam \citep{kingma2014adam} with a starting learning rate of 0.01, with early stopping after ten epochs of no improvement on a held-out validation set as defined by the original datasets. The networks are 2-layer GCN \citep{kipf2016semi}, GAT\citep{velivckovic2017graph} models with hidden dimension set to 128. From this table, training on a randomly generated graph hinders downstream performance, whereas pre-training using a graph with similar characteristics as the original graph results in slight improvements.
\begin{table}[htb]
\renewcommand{\arraystretch}{1.0}
\caption{\label{tab:nc_ec} Comparison of Pre-training followed by fine-tuning for node classification and edge classification tasks. }
\begin{center}
\small
\begin{tabular}{l | l| l | c}
\toprule
\textbf{Dataset} & \textbf{Generator} & \textbf{Model} & \textbf{Accuracy} $\uparrow$\\
\hline
\multirow{3}{*}{\textbf{Cora}}
& \multirow{2}{*}{random}
& GCN & {$0.7470$}  \\\cline{3-4}
& & GAT & {$0.7595$}   \\\cline{2-4}
& \multirow{2}{*}{ours}
& GCN & {$0.7677$}  \\\cline{3-4}
& & GAT & {$0.7720$}   \\ \cline{2-4}
& \multirow{2}{*}{no-pretraining}
& GCN & {$0.76275$}  \\\cline{3-4}
& & GAT & {$0.7650$}   \\\hline
\multirow{3}{*}{\textbf{IEEE-Fraud}}
& \multirow{2}{*}{random}
& GCN & {0.9788} \\\cline{3-4}
& & GAT & {0.9793 }  \\\cline{2-4}
& \multirow{2}{*}{ours}
& GCN & {0.9831} \\\cline{3-4}
& & GAT & {0.9840}  \\ \cline{2-4}
& \multirow{2}{*}{no-pretraining}
& GCN & {0.9823} \\\cline{3-4}
& & GAT & {0.9830}  \\\cline{2-4}
\hline
\end{tabular}
\end{center}
\end{table}

\subsection{When is Graph Structure, Feature and their Alignment Important?}
\label{sec:synth_ablation}
We conducted experiments on a synthetic graph with pre-determined structural and feature properties to answer two questions:

\begin{itemize}
    \item When is essential to align the graph structure with the feature set?
    \item When is graph structure important to begin with?
\end{itemize}
To answer these questions, we considered graphs with high (or low) homophily and high (or low) signal-to-noise ratio (SNR) with respect to the structure and features of the graph. In total, four datasets were generated.

The synthetic graphs were constructed so that the downstream model could learn to discriminate using only the graph structure, the features, or both, depending on the settings of the dataset. The synthetic graphs are generated with 1000 nodes and an edge density of 0.06 (about 24,000 edges). We considered two settings for homophily/SNR: high (0.85/1.5) and low (0.15/0.5). Note that a homophily of $h$ indicates that inter clusters are $h$ times more likely to be connected than intra-clusters. A signal-to-noise ratio (SNR) of SNR indicates how discriminative the features are with the downstream label, where the intra-clusters have the same label. In this experiment, the downstream task was node classification. 

We trained a GAT model that leverages both the graph structure and features with the same configuration as in Section \ref{sec:downstream}. We also trained an XGBoost model that was trained only on the graph features. This is done to be able to delineate the usefulness of the aspects of the graph.

\begin{figure}[htb]
    \centering
    \includegraphics[scale=0.45]{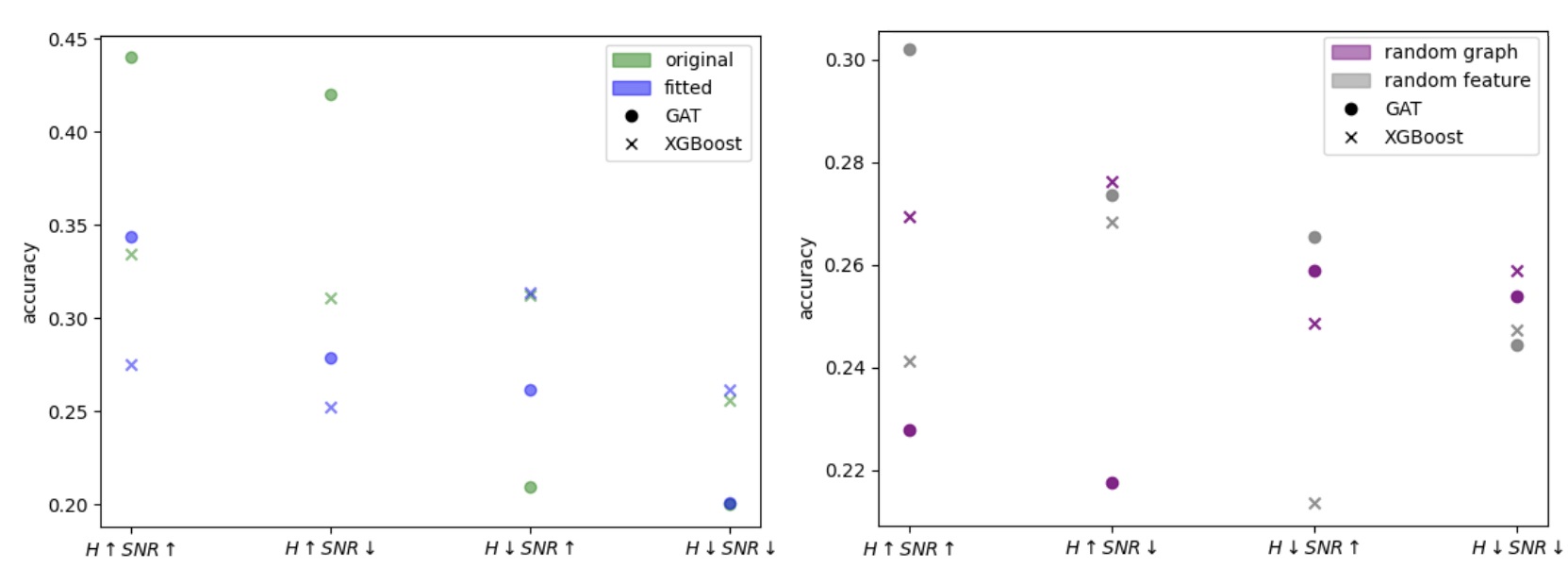}
    \caption{Comparison of training a GAT model on both structure/feature versus training an XGBoost model solely on features for different dataset settings corresponding to high/low homophily ($H\uparrow/\downarrow$) and high/low signal to noise ratio ($SNR\uparrow/\downarrow$). (Left) Depicts the performance on the original synthetic dataset, and fitted corresponding to fitting our proposed generator. (Right) shows the results by replacing the components of the generator with random counterparts.}
    \label{fig:synthablation}
\end{figure}
The results in Figure \ref{fig:synthablation} show that:
\begin{enumerate}
    \item The graph connectivity and graph features are not equally informative for downstream prediction tasks. This can be seen in the figure, where the performance of GAT decreases more when the graph is replaced with a random graph compared to when the features are replaced with random features. For example $H\downarrow SNR \uparrow$ in the right plot, where GAT (circle) in purple denotes when the structure is replaced with random noise, and grey denotes when the features are replaced with random noise.
    \item A noisy graph structure can hinder performance. For example, in the left plot of Figure \ref{fig:synthablation} we can see that XGBoost outperforms GAT on the original synthetic dataset (colored in green) when the graph structure is noisy (i.e., for the cases where $H\downarrow$). In such cases, it may be best to drop the structure and simply use the features and train non geometric models like XGBoost.
    \item Graph alignment is required when both the graph structure and features are informative. However, if the graph structure is not informative, then graph alignment is not necessary, i.e. the randomly aligned graph performs just as well as a fitted graph. 
\end{enumerate}

\subsection{Random Graph Generation Timings}
In Table \ref{tab:random_graph} we presented the experiment with producing big random graphs using the Erdos-Renyi model. We froze the number of nodes and increased the number of edges up to a trillion. We use the same machine with 8 NVIDIA V100 16GB GPUs as in the MAG240m experiments (see Table \ref{tab:mag_scale}).

\begin{table}[h]
\caption{\label{tab:random_graph} Random graph generation timings}
\vspace{-3.0mm}
\begin{center}
\begin{tabular}{c|ccccc}
\toprule
nodes & \multicolumn{5}{c}{100e6}                    \\
edges & 100e9  & 250e9  & 500e9  & 750e9   & 1e12    \\
time  & 22 min & 46 min & 103min & 130 min & 179 min \\ \hline
\end{tabular}
\vspace{-5.0mm}
\end{center}

\end{table}

\subsection{Alignment Comparison}
\label{sec:aligncomparison}
The aligner operates on the graph structure features $F_S: V \rightarrow \Re^{d_{S}}$. The corresponding set of structure features used to correlate with the graph features is not limited to the set used in the paper. As an example, we run an experiment comparing Node2Vec \cite{grover2016node2vec} to the basic statistics used for the aligner in Table \ref{tab:alignment}. As shown for this particular dataset IEEE, using (degrees, pagerank, katz) slightly outperforms Node2Vec.

\begin{table*}[h!]

\caption{\label{tab:alignment} Comparison of the alignment based on different graph structural features and their combinations, averaged over 5 trials. During each trial a single generated graph for all aligners}

\begin{center}
\begin{tabular}{cccc|cc}
\toprule
\multirow{2}{*}{\textbf{Degrees}} & \multirow{2}{*}{\textbf{Pagerank}} & \textbf{Katz}       & \multirow{2}{*}{\textbf{node2vec}} & \multicolumn{2}{c}{\textbf{Degree-Feat Dist-Dist} $\downarrow$} \\
                                  &                                    & \textbf{centrality} &                                    & \textbf{Avg.}       & \textbf{Std.}       \\ \hline
                                  &                                    &                     & X                                  & 0.482               & ± 0.112             \\
X                                 & X                                  & X                   &                                    & 0.479               & ± 0.116             \\
X                                 & X                                  & X                   & X                                  & 0.481               & ± 0.072   \\ \hline   
      
\end{tabular}
\end{center}
\end{table*}

\subsection{Comparison of structural generators}
\label{sec:structgencomparison}
In Table \ref{tab:structural_netgan} the proposed graph structural generator is compared using the same metrics proposed as in \cite{netgan} on the Cora-ML dataset, which contains 2810 nodes and 7981 edges. Random RMAT \cite{rmat} is the baseline random graph generator. In this comparison our method underperforms the NetGan method, though note that our generator is ~4000x faster than NetGan, and supports generating graphs with billions/trillions of edges.

\begin{table*}[!htb]
\renewcommand{\arraystretch}{1.0}
\caption{\label{tab:structural_netgan} Comparison of graph statistics between the CORA-ML graph and graphs generated by various models, averaged over 5 trials. Baseline results are obtained from \cite{netgan}, where * indicates values for the Conf. model that by definition exactly match the original }

\begin{center}
\resizebox{1.0\textwidth}{!}{%

\begin{tabular}{lr|cccccccccccc}
\toprule
\textbf{Graph}  & \textbf{}  & \multicolumn{2}{c}{\textbf{Max.}}   & \multicolumn{2}{c}{\multirow{2}{*}{\textbf{Assortativity}}} & \multicolumn{2}{c}{\textbf{Triangle}} & \multicolumn{2}{c}{\textbf{Power law}} & \multicolumn{2}{c}{\textbf{Clustering}}  & \multicolumn{2}{c}{\textbf{Wedge count}} \\
\textbf{}       & \textbf{}  & \multicolumn{2}{c}{\textbf{degree}} & \multicolumn{2}{c}{}                                        & \multicolumn{2}{c}{\textbf{count}}    & \multicolumn{2}{c}{\textbf{exponent}}  & \multicolumn{2}{c}{\textbf{coefficient}} & \textbf{}           & \textbf{}          \\
                &            & Avg.            & Std.              & Avg.                         & Std.                         & Avg.              & Std.              & Avg.              & Std.               & Avg.               & Std.                & Avg.                & Std.               \\ \hline
CORA-ML         &            & 240             &                   & -0.075                       &                              & 2814              &                   & 1.86              &                    & 2.73e-3            &                     & 101872              &                    \\
Conf. model     &            & *               & *                 & -0.030                       & ± 0.003                      & 322               & ± 31              & *                 & *                  & 3.00e-4            & ± 2.88e-5           & *                   & *                  \\
Conf. model     & (39\% EO)  & *               & *                 & -0.050                       & ± 0.005                      & 420               & ± 14              & *                 & *                  & 4.10e-4            & ± 1.40e-5           & *                   & *                  \\
Conf. model     & (52\% EO)  & *               & *                 & -0.051                       & ± 0.002                      & 626               & ± 19              & *                 & *                  & 6.10e-4            & ± 1.85e-5           & *                   & *                  \\
DC-SBM          & (11\% EO)  & 165             & ± 9.0             & -0.052                       & ± 0.004                      & 1403              & ± 67              & 1.814             & ± 0.008            & 3.30e-3            & ± 2.71e-4           & 73921               & ± 3436             \\
ERGM            & (56\% EO)  & 243             & ± 1.94            & -0.077                       & ± 0.000                      & 2293              & ± 23              & 1.786             & ± 0.003            & 2.17e-3            & ± 5.44e-5           & 98615               & ± 385              \\
BTER            & (2\% EO)   & 199             & ± 13              & 0.033                        & ± 0.008                      & 3060              & ± 114             & 1.787             & ± 0.004            & 4.62e-3            & ± 5.92e-4           & 91813               & ± 3546             \\
VGAE            & (0.3\% EO) & 13.1            & ± 1               & -0.010                       & ± 0.014                      & 14                & ± 3               & 1.674             & ± 0.001            & 1.17e-3            & ± 2.02e-4           & 31290               & ± 178              \\
NetGAN VAL      & (39\% EO)  & 199             & ± 6.7             & -0.060                       & ± 0.004                      & 1410              & ± 30              & 1.773             & ± 0.002            & 2.33e-3            & ± 1.75e-4           & 75724               & ± 1401             \\
NetGAN EO       & (52\% EO)  & 233             & ± 3.6             & -0.066                       & ± 0.003                      & 1588              & ± 59              & 1.793             & ± 0.003            & 2.44e-3            & ± 1.91e-4           & 86763               & ± 1096             \\
Random RMAT     & (0.1\% EO) & 35.0            & ± 5.15            & -0.085                       & ± 0.060                      & 96                & ± 33              & 1.595             & ± 0.008            & 6.13e-4            & ± 1.41e-4           & 120435              & ± 7601             \\
Ours w/o noise  & (0.6\% EO) & 18              & ± 0.7             & -0.038                       & ± 0.006                      & 187               & ± 18              & 1.566             & ± 0.001            & 2.42e-3            & ± 2.39e-4           & 95630               & ± 279              \\
Ours with noise & (0.7\% EO) & 38.4            & ± 4.8             & 0.199                        & ± 0.062                      & 591               & ± 228             & 1.611             & ± 0.011            & 2.93e-3            & ± 6.74e-4           & 134464              & ± 11540           
\end{tabular}
}



\resizebox{1.0\textwidth}{!}{%
\begin{tabular}{lr|cccccccccccc}
\toprule
\textbf{Graph}  & \textbf{}  & \multicolumn{2}{c}{\textbf{Rel. edge}}    & \multicolumn{2}{c}{\textbf{Largest}}    & \multicolumn{2}{c}{\textbf{Claw}}  & \multicolumn{2}{c}{\textbf{Gini}}        & \multicolumn{2}{c}{\textbf{Edge}}    & \multicolumn{2}{c}{\textbf{Characteristic}} \\
\textbf{}       & \textbf{}  & \multicolumn{2}{c}{\textbf{distr. entr.}} & \multicolumn{2}{c}{\textbf{conn. comp}} & \multicolumn{2}{c}{\textbf{count}} & \multicolumn{2}{c}{\textbf{coefficient}} & \multicolumn{2}{c}{\textbf{overlap}} & \multicolumn{2}{c}{\textbf{path length}}    \\
                &            & Avg.                & Std.                & Avg.               & Std.               & Avg.            & Std.             & Avg.               & Std.                & Avg.             & Std.              & Avg.                & Std.                  \\ \hline
CORA-ML         &            & 0.941               &                     & 2810               &                    & 3.1e6           &                  & 0.482              &                     & 1                &                   & 5.61                &                       \\
Conf. model     &            & 0.928               & ± 0.002             & 2785               & ± 4.9              & *               & *                & *                  & *                   & 0.013            & ± 0.001           & 4.38                & ± 0.01                \\
Conf. model     & (39\% EO)  & 0.931               & ± 0.002             & 2793               & ± 2.0              & *               & *                & *                  & *                   & 0.39             & ± 0.0             & 4.41                & ± 0.02                \\
Conf. model     & (52\% EO)  & 0.933               & ± 0.001             & 2793               & ± 6.0              & *               & *                & *                  & *                   & 0.52             & ± 0.0             & 4.46                & ± 0.02                \\
DC-SBM          & (11\% EO)  & 0.934               & ± 0.001             & 2474               & ± 18.9             & 1.2e6           & ± 170045         & 0.523              & ± 0.003             & 0.11             & ± 0.003           & 5.12                & ± 0.04                \\
ERGM            & (56\% EO)  & 0.932               & ± 0.001             & 2489               & ± 11               & 3.1e6           & ± 57092          & 0.517              & ± 0.002             & 0.56             & ± 0.014           & 4.59                & ± 0.02                \\
BTER            & (2\% EO)   & 0.935               & ± 0.000             & 2439               & ± 19               & 2.0e6           & ± 280945         & 0.515              & ± 0.003             & 0.02             & ± 0.001           & 4.59                & ± 0.03                \\
VGAE            & (0.3\% EO) & 0.990               & ± 0.000             & 2810               & ± 0                & 46586           & ± 937            & 0.223              & ± 0.003             & 0.003            & ± 0.001           & 5.28                & ± 0.01                \\
NetGAN VAL      & (39\% EO)  & 0.959               & ± 0.000             & 2809               & ± 1.6              & 1.8e6           & ± 141795         & 0.398              & ± 0.002             & 0.39             & ± 0.004           & 5.17                & ± 0.04                \\
NetGAN EO       & (52\% EO)  & 0.954               & ± 0.001             & 2807               & ± 1.6              & 2.6e6           & ± 103667         & 0.42               & ± 0.003             & 0.52             & ± 0.001           & 5.20                & ± 0.02                \\
Random RMAT     & (0.1\% EO) & 0.975               & ± 0.003             & 3991               & ± 29.8             & 4.6e5           & ± 80875          & 0.355              & ± 0.023             & 0.001            & ± 4e-4            & 4.37                & ± 0.06                \\
Ours w/o noise  & (0.6\% EO) & 0.987               & ± 0.001             & 4071               & ± 6.5              & 2.3e5           & ± 2088           & 0.257              & ± 0.002             & 0.006            & ± 0.001           & 4.59                & ± 0.01                \\
Ours with noise & (0.7\% EO) & 0.969               & ± 0.005             & 3911               & ±  81              & 5.9e5           & ± 128679         & 0.397              & ± 0.027             & 0.007            & ± 0.001           & 4.45                & ± 0.04      \\
\hline
\end{tabular}
}
\end{center}
\end{table*}

\subsection{Degree-Distribution - Feature-Distribution}

\begin{figure}[ht]
    \centering
    \includegraphics[scale=0.32]{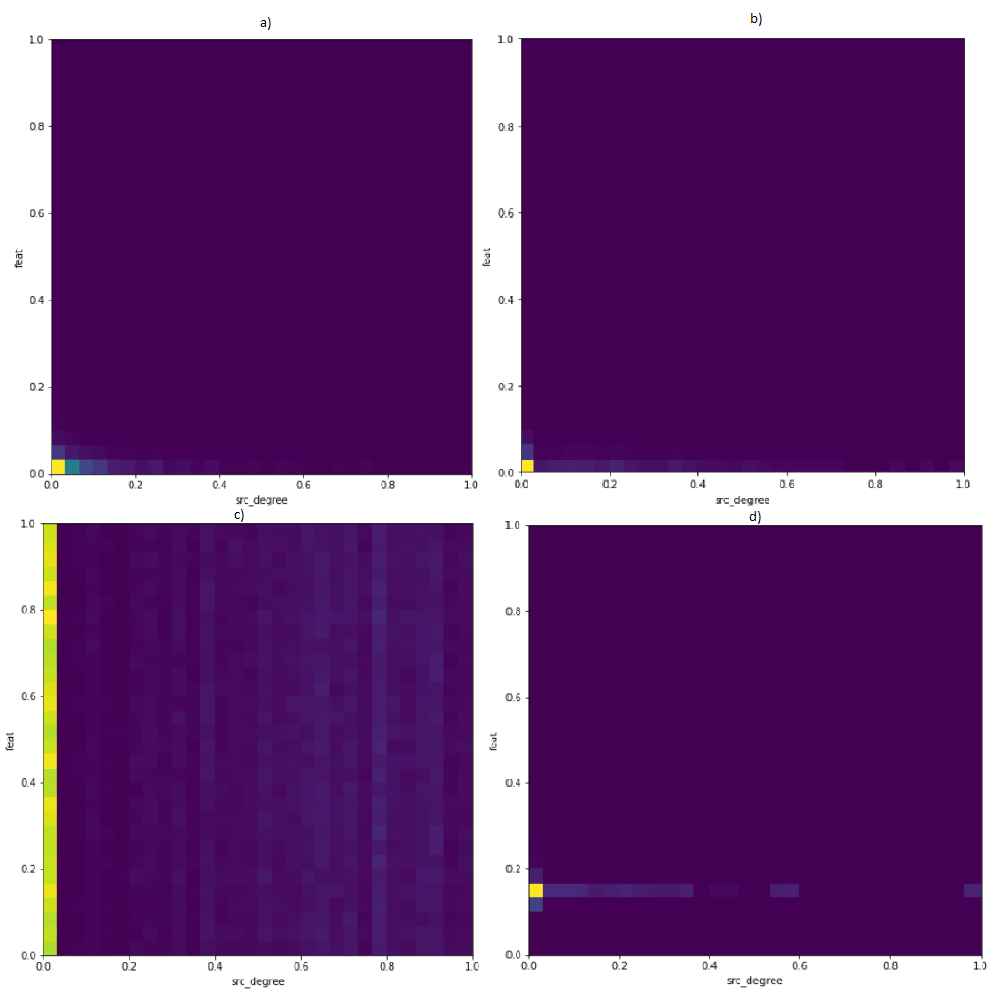}
    \caption{Histograms comparing degree distribution and feature distribution for IEEE-Fraud dataset. a) original graph, b) ours generated, c) randomly generated, d) GraphWorld generated with the added fitting.}
    \label{fig:dd-dist}
\end{figure}

The final graph can be compared visually by plotting the degree distribution vs feature distribution across the feature sets. In Figure \ref{fig:dd-dist} we provide a comparison across the original dataset, our properly generated and the baseline methods. Darker regions in this figure correspond to a lack of feature values for a particular binned degree. Note as the graph is bipartite in this example, the x-axis is the source degree and y axis is the feature distribution. To qualitatively evaluate the aligned features with the graph structure, a plot depicting the degree-distribution versus feature-distribution is used to compare across the methods where visually the heatmaps of the synthetic data should match the original graphs heatmap. This is the case for our proposed method.

\subsection{Comparing Feature CDFS}

\begin{figure}[ht]
    \centering
    \includegraphics[scale=0.09]{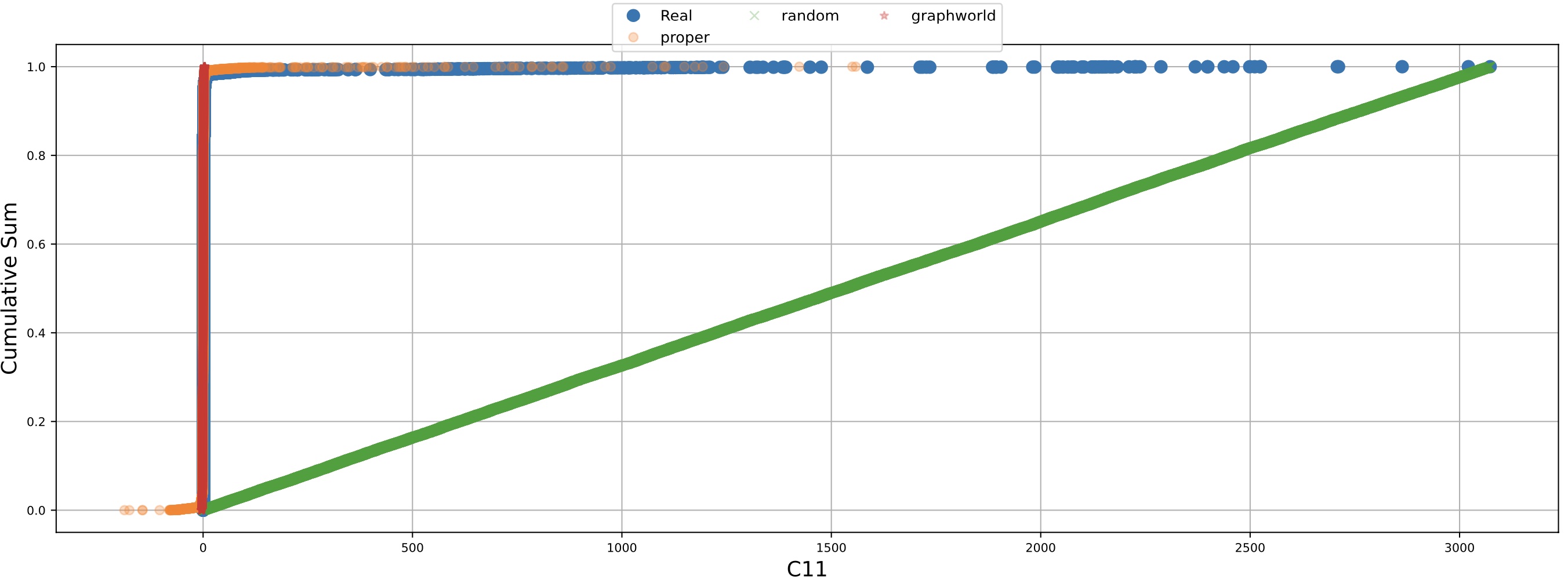}
    \caption{Cumulative distribution comparison on feature column C11 of IEEE-Fraud dataset.}
    \label{fig:cumsum}
\end{figure}

Qualitatively, the cumulative distribution generated using the proposed fitted GAN architecture best resembles the original feature distribution (see Figure. \ref{fig:cumsum}). 

\label{sec:ddetails}
\subsection{Dataset Details}
Constructing graphs from a tabular dataset requires capturing sample relationships. Unlike graph datasets that have the structure and features given directly, sample relations are not immediately given in tabular settings. As a result, these need to be extracted from the data. In practice, tabular datasets are from very diverse domains ranging from fraud detection to recommender systems and electronic health records; these relationships require to be inferred from the features with domain knowledge. The feature columns used for extracting the corresponding relationship and edges from the tabular dataset are summarized in Table. \ref{tab:data_construction}.

\begin{table}[htb]
\renewcommand{\arraystretch}{1.0}
\caption{\label{tab:data_construction} Details on how to construct a graph from tabular features. The node column corresponds to the set of features used to construct node types, and the condition column details the condition that must be satisfied for an edge to exist between the nodes. }
\begin{center}
\small
\begin{tabular}{ccc}
\toprule
 \thead{Dataset} & \thead{Nodes} & \thead{Condition} \\
 \hline
 \makecell{Tabformer} & \makecell{concat(User, Card) \\ Merchant ID} & \makecell{same row}\\
 \hline
 \makecell{IEEE-Fraud} & \makecell{concat($7$ features) \footnotemark \\ concat($2$ features) \footnotemark } & \makecell{same row}\\
 \hline
 \makecell{Paysim} &  \makecell{nameOrig\\ nameDest} & \makecell{same row}\\
 \hline
 \makecell{Credit} & \makecell{concat(first, last) \\ merchant} & \makecell{same row}\\
 \hline
 \makecell{Home-Credit} & \makecell{id} & \makecell{same concat($8$ features) \footnotemark}\\
 \hline
\makecell{Travel-Insurance} & \makecell{id} & \makecell{AnnualIncome \\ Age}\\
\hline
\end{tabular}
\end{center}
\end{table}

\addtocounter{footnote}{-3} 
\stepcounter{footnote}\footnotetext{(ProductCD, R\_emaildomain)}
\stepcounter{footnote}\footnotetext{(addr1, addr2, card1, card2, card3, card4, card5, card6)}
\stepcounter{footnote}\footnotetext{(OWNERSHIP\_TYPE, P1\_SEX, POL\_STATUS, CONTENTS\_COVER, SUBSIDENCE, SEC\_DISC\_REQ, MTA\_FLAG, P1\_EMP\_STATUS)}
\subsection{Comparing Degree Distribution}

Our structure generator fits its parameters to the degree distribution of the original graph according to \eqref{eq:gstructure_generator_loss}. Estimating the quality of that fitting can be done visually by comparing plots like in Figure \ref{fig:ddhop}. Still, in most cases, it isn't easy to assess whether the improvement of the fitting procedure makes a degree distribution of the synthetic graph closer to the original. This holds especially for comparing a different size synthetic graph in terms of $N$,$M$, and $E$ than the original graph. 
We propose a single scalar metric that captures the alignment of degree distribution for two graphs. This metric is calculated as follows:

\begin{equation}\label{eq:cdd}
   DCC=\frac{1}{K} \sum_{k \in logspace(0,1)}{ 
   \frac{
        c_k^{norm} - \hat{c_k^{norm}}
        }{c_k^{norm}}
   }
\end{equation}

where $K$ is the number of distinct degree $k$ sampled logarithmically from $[0,1]$ 
$c_k^{norm}$, $\hat{c_k^{norm}}$ are normalized degree distributions i.e. degree is normalized by the maximum degree in the graph and the number of nodes is normalized by maximum $c_k$ and $\hat{c_k}$ respectively.

Normalization of the degree distribution is needed as $G$ and $\hat{G}$ may have different sizes both in terms of the number of nodes and edges. For $G$ and $\hat{G}$ of the same number of edges $E$ the $DCC$ in \eqref{eq:cdd} simplifies to

\begin{equation}\label{eq:cdd_simple}
   DCC=\frac{1}{K} \sum_{k \in logspace(0,k_{max})}{ 
   \frac{
        c_k - \hat{c_k}
        }{c_k}
   }.
\end{equation}

For generating larger graphs, we need to ensure that graph density is preserved i.e.:

\begin{equation}\label{eq:graphden}
\frac{E}{N*M} = \frac{\hat{E}}{\hat{N}*\hat{M}},
\end{equation}

where $\hat{E} , \hat{N}, and \hat{M}$ are number of edges and nodes in partites of $\hat{G}$.

For example, for a homogeneous graph when increasing the number of nodes twice, one needs to increase the number of edges four times to preserve the constant density.

\begin{figure}[htb]
\begin{tabular}{cc}
  \includegraphics[width=65mm]{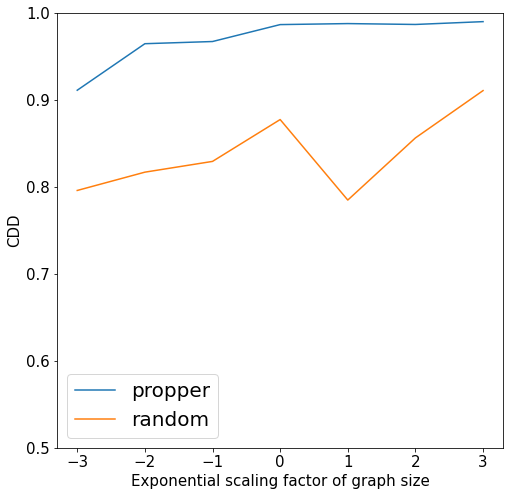} (a)
  \includegraphics[width=65mm]{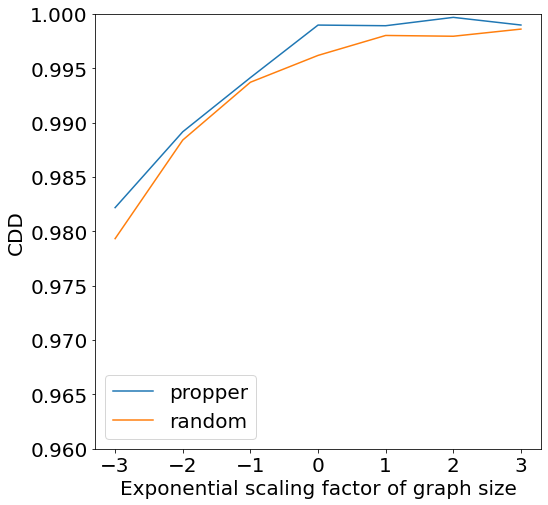} (b)  \\[6pt]
\end{tabular}
\caption{\label{fig:cdd} CDD coefficient calculated according to \ref{eq:cdd} for different datasets (a) Tabformer, (b) IEEE-Fraud. Two generating models are compared: our method marked as 'propper' and random which is ER \cite{erdos59a}. The X-axis of each graph is an exponential scaling factor by which the number of nodes in each partite is multiplied e.g. $0$ means graph of the same size, +3 means graph for which $\hat{N} = 2^{3} N$,$\hat{M} = 2^3 M$ and $\hat{E} = 2^6 E$, -3 means graph for which $\hat{N} = 2^{-3} N$,$\hat{M} = 2^{-3} M$ and $\hat{E} = 2^{-6} E$, etc}
\end{figure}

Our method consistently outperforms ER model not only for the same graph size but across all scaling factors, see Figure \ref{fig:cdd}. It also provides very high values of CDD for large graphs which in fact means that when generating large synthetic graphs, the degree distribution curve shape preserves its power-law shape. This statement holds for generating smaller graphs that is equivalent to subsampling a larger graph in a stratified way where nodes' degree proportion is preserved - this can be primarily seen in the Tabformer dataset where partites are imbalanced in size ($2^{17}$ user nodes and $2^{11}$ merchant nodes). 


\section{Adding Noise to Structure Generator}
Graph $\hat{G}$ generated by \eqref{eq:theta_matrix_eq} will produce oscillations on the degree distribution as described in \cite{skg_hitch}. To address oscillations we propose to add a noise component on each $max(m,n)$ step of \eqref{eq:theta_matrix_eq}. This changes \eqref{eq:theta_matrix_eq} to


\begin{equation} \label{eq:theta_matrix_eq_noise}
    \theta = \underbrace{\theta_{S,0} \otimes ... \otimes \theta_{S,min(n,m)}}_{min(m,n) times}  \otimes \underbrace{\theta_{H,0} \otimes ... \otimes \theta_{H,min(0,n-m)}}_{min(0,n-m) times}  \otimes \\
    \underbrace{\theta_{V,0} \otimes  ... \otimes \theta_{V,min(0,m-n)}}_{min(0,m-n) times},
\end{equation}

where $\theta_{S,i}$, $\theta_{H,i}$, $\theta_{V,i}$ are noisy versions of $\theta_{S}$, $\theta_{H}$, $\theta_{V}$ from \eqref{eq:theta_matrix_eq}, respectively. If noise is not added then \eqref{eq:theta_matrix_eq_noise} and \eqref{eq:theta_matrix_eq} are equivalent. $\theta_{S,i}$ (and analogically $\theta_{H,i}$, $\theta_{V,i}$) are modifications of $\theta_{S}$ (and $\theta_{H}$, $\theta_{V}$, respectively)

\begin{equation} \label{eq:noise_alone}
    \theta_{S,i} = \theta_{S} + N_i(\theta_{S}),
\end{equation}

analogically for  $\theta_{H,i}$, $\theta_{V,i}$. Mean value of noise added to cascade \eqref{eq:theta_matrix_eq} has to be zero, but careful mathematical analysis  shows that also elements of noise matrix $N_i$ added to matrix $\theta_S$ (and respectively $\theta_{H}$, $\theta_{V}$) have to be zero.

Noise added by $N_i$ depends on $\theta_{S}$ and in practice can be controlled by a single parameter sampled from uniform distribution. An exemplary noise for symmetric $\theta_{S}$ can be

\begin{equation} \label{eq:noise_sample}
 N_i = \begin{bmatrix}
                \frac{-2n_f*a}{a+d} & nf \\
                nf & \frac{2n_f*a}{a+d}
            \end{bmatrix}
            \linebreak
n_f \sim U [min(\frac{a+d}{2},b,c)],
\end{equation}

Where $U[x,y]$ denotes the uniform distribution .... Overall, adding noise on each step of the generator requires only $max(n,m)$ parameters.

\section{Chunked Generation}
\label{sec:chunked_generation}
The R-MAT algorithm operates by recursively subdividing the adjacency matrix $A$. The $i$-th subdivision corresponds to the single matrix $\theta_i \in \Re^{2\times2}$ and may be interpreted as choosing a bit for the source and destination node of the sampled edge.
For the case of generating large graphs, producing a graph $\hat{G}$ by sampling $E$ edges from the distribution $\theta$ may not fit into the available memory. In order to parallelize generation and decrease memory consumption, $\theta$ is represented as $\theta_{pref}\otimes\theta_{gen}$, where $\theta_{pref}$ is used to generate a unique chunk prefix to avoid id-overlap and both terms have the form $\theta_S^{\otimes n_x}$ as in eq. \ref{eq:theta_matrix_eq}. 
As each edge is sampled independently, we can replace sampling prefixes from $\theta_{pref}$ by the expected value of the edges for the given prefix $E_{pref}=E\cdot\mathbb{E}[\theta_{pref}]$. To this end, to produce the $i$-th chunk we sample $E^{i}_{pref}$ edges from $\theta_{gen}$ and prepend the $i$-th prefix to them. The prefixes guarantee us that there will be no edge overlap between chunks and the final graph is simply constructed by concatenating them to obtain the graph $\hat{G}$.

\section{Graph Generation Comparison}

\begin{figure}
    \centering
    \includegraphics[scale=0.25]{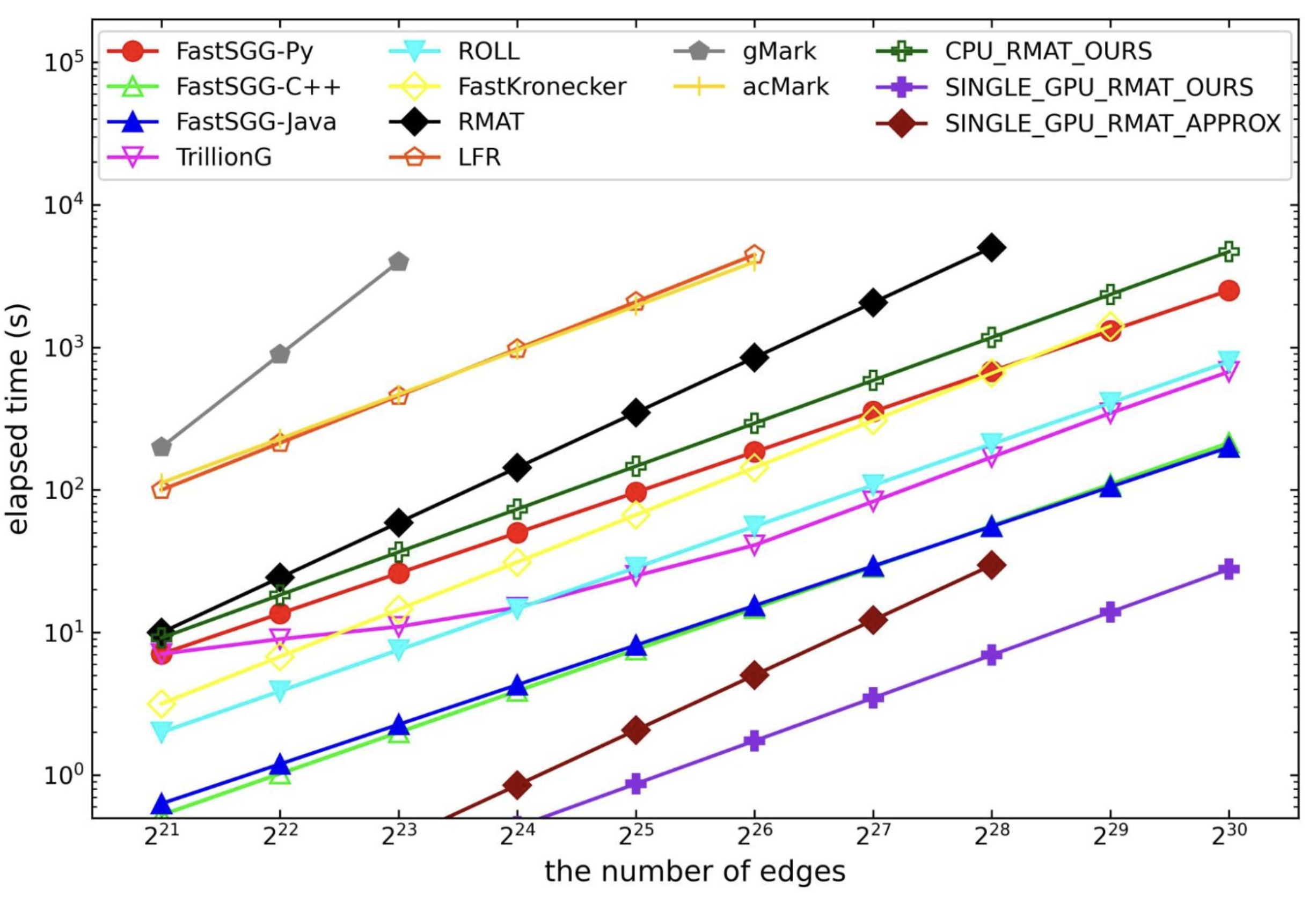}
    \caption{Graph generator throughput comparison.}
    \label{fig:speed}
\end{figure}

In Figure \ref{fig:speed}, we provide a comparison with other large scale graph generators such as FastSGG \cite{wang2021fastsgg}, TrillionG \cite{park2017trilliong}, FastKronecker \cite{leskovec2010kronecker}. The results for these generators were taken from \cite{wang2021fastsgg}, which uses a machine with Intel Xeon E5-2630 CPU (2.20GHz), and we replicate this setup with our RMAT generator implementation both on CPU and GPU. On CPU the machine used a single Intel(R) Xeon(R) CPU E5-2698 v4 @ (2.20GHz), which is shown as dark green in the figure. Our GPU implementation run on single V100 16GB GPU, outperforms all the other generators by a large margine. Note that brown is extrapolated by taking the dark RMAT curve implementation and show how it would scale if we replace it with GPU implementation (assuming their implementation has the same performance as ours). This is done as we use a slightly different CPU. In this case comparing the brown curve with the other generators we still outperform other methods.

\section{Experiment Details}
\label{sec:exp_app}
We provide details on hyperparameters used in our experiments. Our structural generator has a fitting portion that fits the underlying dataset without requiring the user to specify the parameters. The XGBoost\footnote{\href{https://rapids.ai/xgboost.html}{https://rapids.ai/xgboost.html}} aligner learning rate is set to a default value of 0.1, max depth of 5, and the number of estimators is set to 100, with an alpha value of 10., We set the hidden dimensions to be equal to the input dimension. For all experiments for training to our proposed method, we use Adam \citep{kingma2014adam} optimizer with an initial learning rate of $1e-3$ decayed every ten epochs by a factor of $0.1$ and trained for a maximum of $20$ epochs with early stopping. Note that we train on the complete input data and do not split the dataset into the train, validation, and test splits as we aim to generate a single graph while our input is also a single graph. For most datasets, it suffices to train the feature generator for about $5$ epochs. Note that for datasets that contain categorical columns, the embedding size for these columns is set to $min(600, round(1.6 * |D|^{0.56})$ where $|D|$ is the number of unique values for the categorical column. Our code is available on Github \footnote{\href{https://github.com/<anon>/<anon>}{https://github.com/<anon>/<anon>}}

\section{Software \& Hardware}
\label{sec:software}
Experiments were run on a machine with 8 16GB V100s, 512GB RAM, and 80 CPUs Intel XEON E5-2698 v4 @ 2.20GHz using CUDA version 11.8.

\begin{table}[h]%
    \renewcommand{\arraystretch}{1.3}
    \caption{Python dependencies.}

        \begin{center}
            \texttt{
                \begin{tabular}{l|l}
                    \toprule
                        \textbf{Dependency} & \textbf{Version} \\
                        \hline
                        python & 3.8.10 \\
                        numpy & 1.22.2 \\
                        scipy & 1.7.0 \\
                        pandas & 1.5.2 \\
                        scikit-learn & 0.24.2 \\
                        xgboost & 1.6.2 \\
                        pytorch & 1.14.0 \\
                        dgl & 0.9.1 \\
                        cupy &  11.0.0 \\
                        cudf & 21.12 \\
                    \bottomrule
                \end{tabular}
            }
        \end{center}
    \label{tab:swdeps}
\end{table}%


\end{document}